\pdfoutput=1

\documentclass[11pt]{article}

\usepackage{acl}

\usepackage{times}
\usepackage{latexsym}

\usepackage[T1]{fontenc}

\usepackage[utf8]{inputenc}

\usepackage{microtype}

\usepackage{inconsolata}

\usepackage{array}
\usepackage{multirow}
\usepackage{booktabs}
\usepackage{xcolor}

\usepackage{graphicx}
\usepackage{subcaption}

\usepackage{soul}
\usepackage{nicefrac}
\usepackage{amsfonts}
\usepackage{amsmath}
\usepackage{amsthm}
\usepackage{algorithm}
\usepackage{algorithmic}

\usepackage[capitalise]{cleveref}

\newcolumntype{C}[1]{>{\centering\arraybackslash}m{#1}}

\newtheorem{definition}{Definition}
\newtheorem{theorem}{Theorem}
\newtheorem{lemma}{Lemma}

\DeclareMathOperator*{\argmin}{arg\,min}

\newcommand{\eg}{\emph{e.g.}}
\newcommand{\ie}{\emph{i.e.}}
\newcommand{\wrt}{\emph{w.r.t.}\,}

\newcommand{\alphabar}{\bar{\alpha}}
\newcommand{\betabar}{\bar{\beta}}
\newcommand{\ep}{\boldsymbol{\varepsilon}}

\newcommand{\loss}{\mathcal{L}}
\newcommand{\lossrm}[1]{\loss_\mathrm{#1}}
\newcommand{\lvlb}{\lossrm{vlb}}
\newcommand{\lround}{\lossrm{round}}
\newcommand{\lvanilla}{\lossrm{text}}
\newcommand{\lanchor}{\lossrm{anchor}}
\newcommand{\lours}{\lossrm{}}

\newcommand{\e}{\boldsymbol{e}}
\newcommand{\efun}{\e_\phi}
\newcommand{\emat}{\e_\phi}
\newcommand{\eset}{\e_\phi}

\newcommand{\x}{\boldsymbol{x}}
\newcommand{\xhid}{\mathbf{x}}
\newcommand{\y}{\boldsymbol{y}}

\newcommand{\z}{\mathbf{z}}
\newcommand{\zhat}{\hat{\z}_0}
\newcommand{\zvec}{\boldsymbol{z}}

\newcommand{\ani}{\mathrm{ANI}}

\newcommand{\fdg}{f_\mathrm{dg}}
\newcommand{\dgs}{\mathrm{DGS}}
\newcommand{\dgsmax}{\dgs_\mathrm{MAX}}

\newcommand{\limcond}{\substack{\betabar \to 0 \\ d \to \infty}}
\newcommand{\convergesto}{\xrightarrow[\limcond]{P}}

\title{Empowering Diffusion Models on the Embedding Space for Text Generation}

\author{
  Zhujin Gao$^{1,2}$\thanks{Equal contribution.}, 
  Junliang Guo$^3$\footnotemark[1],
  Xu Tan$^3$, Yongxin Zhu$^{1,2}$, Fang Zhang$^{1,2}$, \\
  {\bf Jiang Bian$^3$, Linli Xu$^{1,2}$\thanks{Corresponding author.}} \\ 
  $^1$School of Computer Science and Technology, University of Science and Technology of China \\
  $^2$State Key Laboratory of Cognitive Intelligence \\
  $^3$Microsoft Research Asia \\
  \texttt{gaozhujin@mail.ustc.edu.cn},\;
  \texttt{\{junliangguo, xuta\}@microsoft.com}\\
  \texttt{\{zyx2016, fangzhang\}@mail.ustc.edu.cn},\;
  \texttt{jiabia@microsoft.com}\\
  \texttt{linlixu@ustc.edu.cn}\\
}

\begin{document}
\maketitle

\begin{abstract}
Diffusion models have achieved state-of-the-art synthesis quality on both visual and audio tasks, and recent works further adapt them to textual data by diffusing on the embedding space.
In this paper, we conduct systematic studies of the optimization challenges encountered with both the embedding space and the denoising model, which have not been carefully explored.
Firstly, the data distribution is learnable for embeddings, which may lead to the collapse of the embedding space and unstable training. To alleviate this problem, we propose a new objective called the anchor loss which is more efficient than previous methods.
Secondly, we find the noise levels of conventional schedules are insufficient for training a desirable denoising model while introducing varying degrees of degeneration in consequence. To address this challenge, we propose a novel framework called noise rescaling.
Based on the above analysis, we propose Difformer, an embedding diffusion model based on Transformer. Experiments on varieties of seminal text generation tasks show the effectiveness of the proposed methods and the superiority of Difformer over previous state-of-the-art embedding diffusion baselines.\footnote{Code is available at \url{https://github.com/zhjgao/difformer}}
\end{abstract}

\section{Introduction}

A wave of diffusion models~\citep{sohl2015deep, ho2020denoising, song2020score} 
is sweeping the generation tasks~(\eg, image and audio synthesis) recently, showing their great capacity for high-quality data generation. Diffusion models are a family of iterative generative models, which are trained to recover corrupted data and then generate data by gradually refining samples from the pure noise. This procedure enables the model to make subtle refinements of output samples in a multi-step denoising process, and thus generate high-fidelity and diverse samples~\citep{dhariwal2021diffusion, nichol2021improved, ho2021classifier, rombach2022high, chen2020wavegrad, kong2020diffwave}.

The booming achievements in vision and audio domains inspire researchers to delve into the realm of text generation.
Diffusion models introduce a novel noising paradigm and a training objective other than token prediction, establishing an alternative form of language models, which exhibits the potential to foster an enhanced comprehension of language modeling.
From a higher perspective, this investigation generalizes the diffusion model across modalities, and further contributes to a unified multimodal framework~\citep{bao2023one, tang2024any}.
Nonetheless, the exploration is still at an initial stage. Recent works~\citep{li2022diffusion, gong2022diffuseq, strudel2022selfconditioned} basically convert the discrete tokens to embeddings and then utilize continuous diffusion models to generate them, which can be termed embedding diffusion models. 
These preliminary attempts follow the original model to deal with the embeddings, with little consideration of the unique properties and the optimization challenges of the embedding space and the denoising model.
 
In this paper, we explore the embedding diffusion model from two perspectives separately, \ie, the embedding space and the denoising model, based on which we conduct a thorough study respectively.
Firstly, for diffusion models on image and audio generation, the ground truth data is stationary during training. In contrast, it is learnable for the textual data (\ie, embeddings), which may cause the collapse of the embedding space and introduce instability to the training of the model. 
To avoid the collapse caused by dynamically shifting embedding parameters, we propose an anchor loss function to attain well-distributed embeddings and stabilize the training process. The detailed analysis is presented in \cref{sec:collapse}.

Secondly, in \cref{sec:degeneration}, we find that in the high dimensional embedding space, the insufficient noise results in a simple denoising task, which causes the degeneration of the model.
To tackle this challenge, we propose a novel framework named noise rescaling, which is orthogonal to the choice of the noise schedule and applicable to any existing schedules. Specifically, we define an index termed degeneration score as a measurement of the degree of degeneration. Guided by the degeneration score, we can apply a noise rescaling procedure to prevent the model from degenerating.

Based on the above discussion, we propose an integrated framework of Difformer, a denoising diffusion Transformer model.
We conduct experiments on a variety of important text generation tasks including machine translation, text summarization, paraphrasing, text simplification, and question generation. On these benchmark datasets, Difformer outperforms diffusion-based and iteration-based non-autoregressive baselines and achieves state-of-the-art performance among embedding diffusion models.
Further experiments demonstrate the superiority of Difformer over baselines including LLMs in quality, diversity, and efficiency, emphasizing the potential of diffusion models for text generation in the era of LLMs.

\section{Background}

\paragraph{Diffusion Models}

Denoising diffusion probabilistic models~\citep{sohl2015deep,ho2020denoising} 
utilize a forward process to perturb the data with Gaussian noise, and a reverse process to restore the data symmetrically. \citet{ho2020denoising} develop the approach by specific parameterizations, achieving comparable sample quality with state-of-the-art generative models such as GANs~\citep{goodfellow2014generative}. After that, great improvements have been made by many following works~\citep{song2020denoising, dhariwal2021diffusion, nichol2021improved, rombach2022high} both in quality and efficiency.
Given a data sample $\z_0 \in \mathbb{R}^{d}$, the denoising diffusion probabilistic model gradually perturbs it into a pure Gaussian noise $\z_T \sim \mathcal{N}(\mathbf{0}, \mathbf{I})$ through a series of latent variables $\z_1, \cdots, \z_T$ in the forward process:
\begin{equation*}
    q(\z_t | \z_0) = \mathcal{N} \left(
        \z_t;
        \sqrt{\alphabar_t} \z_0,
        \betabar_t \mathbf{I}
    \right),
\end{equation*}
where $\alphabar_t, \betabar_t$ are hyper-parameters controlling the noise level added at timestep $t$, which form the noise schedule. Usually, these hyper-parameters are set to satisfy $\alphabar_t := \prod_{i = 0}^t \alpha_i, \alpha_t + \beta_t = 1,$ and $ \alphabar_t + \betabar_t = 1$.
The reverse process is parameterized as:
\begin{equation*}
    p_\theta(\z_{t - 1} | \z_t) = \mathcal{N} \left(
        \z_{t - 1};
        \boldsymbol{\mu}_{\theta}(\z_t, t),
        \mathbf{\Sigma}_\theta(\z_t, t)
    \right),
\end{equation*}
where $\boldsymbol{\mu}_\theta(\cdot)$ and $\mathbf{\Sigma}_\theta(\cdot)$ are the predicted mean and covariance of $q(\z_{t - 1} | \z_t)$, and $\theta$ denotes the model parameters.
After parameterization, we utilize a simplified variational lower-bound as the objective function
\begin{equation}
    \label{equ:vlb}
    \lvlb = \mathbb{E}_{\z_0, \z_t, t} \left [
        \lVert \zhat(\z_t, t) - \z_0 \rVert^2
    \right ],
\end{equation}
where $\zhat(\z_t, t)$ is the model prediction of the original data $\z_0$ given $\z_t$. The detailed derivation can be found in \cref{sec:derivation}.

\begin{figure*}[t]
    \centering
    \includegraphics[width=0.9\textwidth]{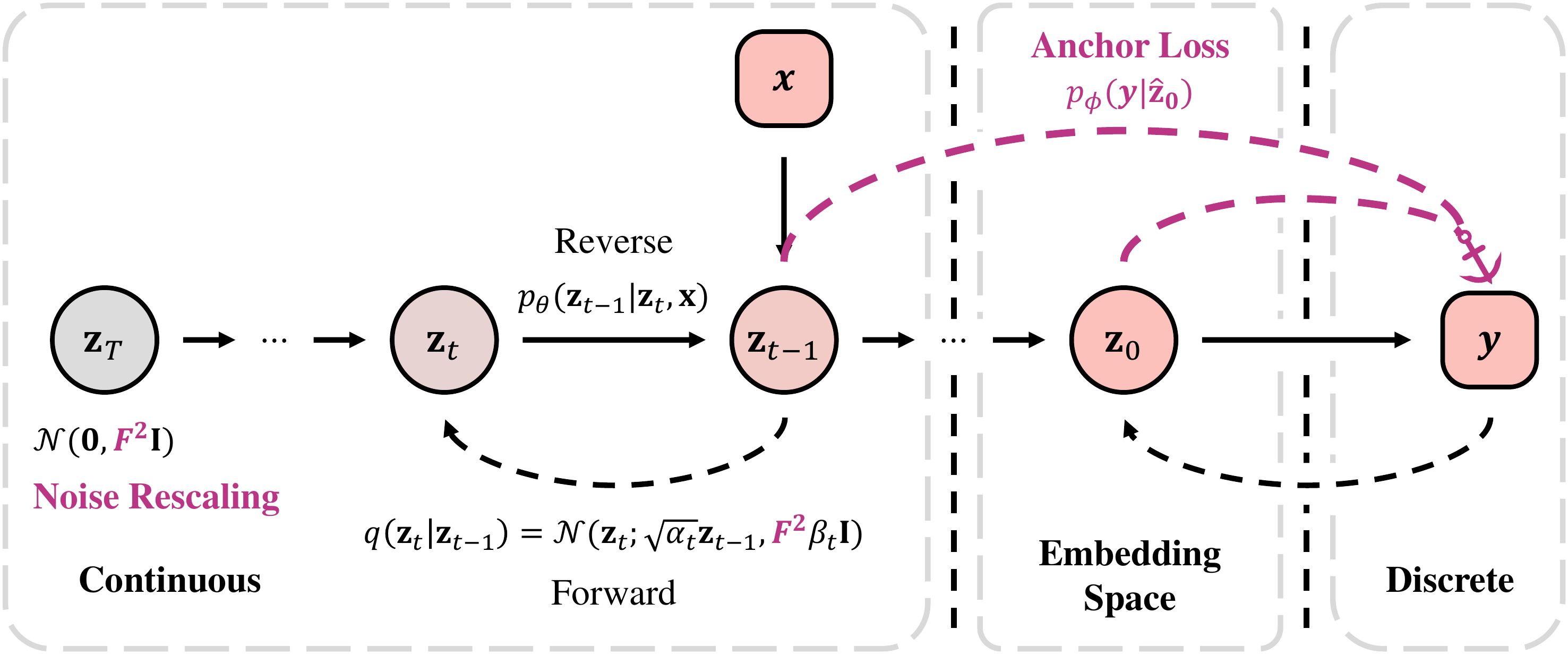}
    \caption{An overview of the Difformer, including the proposed techniques, \ie, the anchor loss, and the noise rescaling.}
    \label{fig:overview}
\end{figure*}

\paragraph{Diffusion Models for Text Generation}

The breakthrough of diffusion models on continuous data encourages people to explore their potential on discrete textual data.
The definition of forward and reverse processes is the key question for diffusion models. Recent works mainly follow two directions.

Firstly, discrete diffusion models on categorical distributions are proposed~\citep{hoogeboom2021argmax, austin2021structured, savinov2021step, reid2022diffuser}, by which sentences are corrupted and refined at the token level. However, these kinds of corruption are coarse-grained.
Attempts have been made to explore modeling on surrogate representations of discrete data such as analog bits~\citep{chen2022analog} and simplex~\citep{han2023ssd}. Nevertheless, these representations carry little semantic information about tokens, which implies that the distances in this space can not accurately reflect semantic correlations between tokens.

In contrast, embedding diffusion models~\citep{li2022diffusion,strudel2022selfconditioned, gong2022diffuseq, ye2023dinoiser} introduce an additional embedding step and rounding step in the forward and reverse processes respectively. The embedding step converts tokens into learnable or pre-trained embeddings, which carry semantic information, and then a continuous diffusion process is able to add Gaussian noise to these embeddings, achieving a fine-grained noising procedure. 
Mathematically, given a sequence of tokens $\y = [y_1, y_2, \cdots, y_n]$, the embedding step can be denoted as $\z_0 \sim \mathcal{N} (\efun(\y), \beta_0 \mathbf{I})$ where $\efun(\cdot)$ denotes the embedding lookup function.
The rounding step turns predicted embeddings back to discrete tokens, which can be expressed as a softmax distribution over the vocabulary $p_{\phi}(\y| \z_0)$, and is trained by an extra loss function $\lround = \mathbb{E}_{\y, \z_0} [-\log p_\phi(\y | \z_0)]$. The parameters of this step and the embedding step are tied. The final loss function is written as:
\begin{equation*}
    \lvanilla = \lvlb + \lround.
\end{equation*}

Nevertheless, these works directly adapt continuous diffusion models to embeddings, without considering the gap between the learnable embedding space and the stationary image or audio data, as well as the distinctive requirements of the denoising model established on the embedding space.

\section{Methodology}

This section elucidates the challenges inherent in optimizing embedding diffusion models and presents our corresponding solutions. We start with an introduction to the model architecture.
The model architecture is based on Transformer~\citep{vaswani2017attention}, which consists of an encoder and a decoder.
The decoder, as the main stem component, is considered as two separate parts in this paper, namely the embeddings $\emat = [\e_1, \e_2, \cdots, \e_V] \in \mathbb{R}^{d \times V}, \efun(\y) = [\e_{y_1}, \e_{y_2}, \cdots, \e_{y_n}]$ 
and the denoising model $f_\theta(\cdot)$, which denotes the stacked decoder layers. Notably, this paper define $\z_0 = \efun(\y)$.
The encoder provides the representation $\xhid = \mathrm{Encoder}(\x)$ of the condition sentence $\x = [ x_1, x_2, \cdots, x_m]$.

\subsection{Collapse of the Embedding Space}
\label{sec:collapse}

\paragraph{Analysis of the Collapse Problem}

The data space is usually fixed for continuous data (\eg, image and audio), while it is learned from scratch for discrete textual data~(\ie, embeddings), which therefore shifts dynamically during training.
Original diffusion models rely on the loss function \cref{equ:vlb} to learn to estimate the clean data sample $\z_0$.
Nevertheless, when directly adapting this objective to the embedding diffusion model, the embedding space will collapse.
As a result, the embeddings of different tokens will be less distinguishable and non-uniformly distributed in the space, which considerably limits the representation capacity and quality of the embeddings. On the contrary, the model could achieve better performance with more isotropic embeddings~\citep{gao2018representation, li2020sentence}.

Recent works of diffusion on embeddings~\citep{li2022diffusion, gong2022diffuseq} introduce the rounding loss $\lround$ from the derivation of the variational lower bound, which discriminates the correct embeddings from others given their noised counterparts, therefore enforces the embeddings are distinguishable and informative, alleviating the collapse objectively. We could regard this additional loss function as a regularization term for the embeddings.
Nonetheless, only a minor level of perturbation is involved from $\y$ to $\z_0$, thereby the rounding loss is only able to apply a relatively weak constraint on the embeddings.

Our empirical evidence also corroborates the limitation of the rounding loss. We observe that the rounding loss undergoes a steep descent and falls to near zero in the initial stages of training, which implies the rounding loss can be effortlessly addressed and fails to conduct strong enough regularization to the embeddings. Therefore, the embedding space is undesirable and eventually leads to unsatisfactory performance.
Concurrently, the instability in training also emerges as a problem during training. Even if careful tuning of the hyper-parameters is performed to relieve anisotropy, the performance is still inferior.

\paragraph{Anchor Loss}

To emphasize the effect of the regularization term, we propose a training objective named the anchor loss
\begin{equation*}
    \lanchor = \mathbb{E}_{(\x, \y), \z_t, t} [-\log p_\phi(\y | \zhat(\z_t, \xhid, t))].
\end{equation*}
Compared with $\lround$, $\lanchor$ utilizes the model prediction of $\z_0$ as the input, which involves a large discrepancy with $\z_0$ due to the prediction error of the denoising model. Consequently, to ensure these highly noisy representations are identified as the correct tokens, the anchor loss employs a stronger regularization to the embeddings to prevent collapse.
Additionally, besides $\lvlb$, the anchor loss creates another pathway between the denoising model and the target sentences, through which the model could receive feedback from the ground truth, maintaining the training stability.
Finally, our training objective is written as
\begin{equation}
    \label{equ:ours}
    \lours = \lvlb + \lanchor.
\end{equation}

Empirically, we use self similarity~\citep{ethayarajh2019contextual} as the \textit{anisotropy score} to measure the severity of collapse:
\begin{equation*}
    \ani = \frac{1}{V(V - 1)} \sum_{i=1}^V \sum_{j=1, j \neq i}^V \cos (\e_i, \e_j).
\end{equation*}
Essentially, the higher the anisotropy score is, the more severe the collapse is. The anisotropy score as well as the performance obtained by each loss function can be found in \cref{tab:ani}. With only $\lvlb$ or $\lvanilla$, the anisotropy score demonstrates that the embeddings are non-uniformly distributed, resulting in unsatisfactory results. On the contrary, the embeddings are well-distributed across the entire space with the anchor loss, and thus the model reaches competitive performance~(in BLEU~\citep{papineni2002bleu}).
Alternatively, utilizing pre-trained embeddings and freezing them during training could also avoid collapse. As shown in the experimental results, the frozen embeddings alleviate the collapse remarkably, however, they are suboptimal for the problem. Detailed discussion can be found in \cref{sec:frozen_emb}.

\begin{table}[t]
    \small
    \centering
    
    \begin{tabular}{lcc}
        \toprule
        \textbf{Loss} & $\ani$ & BLEU \\
        \midrule

        $\lvlb$ & $0.99$ & $0.07$ \\
        $\lvanilla$ & $0.32$ & $27.89$ \\
        $\lours$ & $0.03$ & $34.48$ \\
        \bottomrule
    \end{tabular}
    \caption{The anisotropy score and performance of each loss function on the IWSLT14 De-En dataset with \emph{linear} schedule.}
    \label{tab:ani}
\end{table}

\begin{figure*}[t]
    \centering
    \hspace*{\fill}%
    \begin{subfigure}{0.3\textwidth}
        \includegraphics[width=\textwidth]{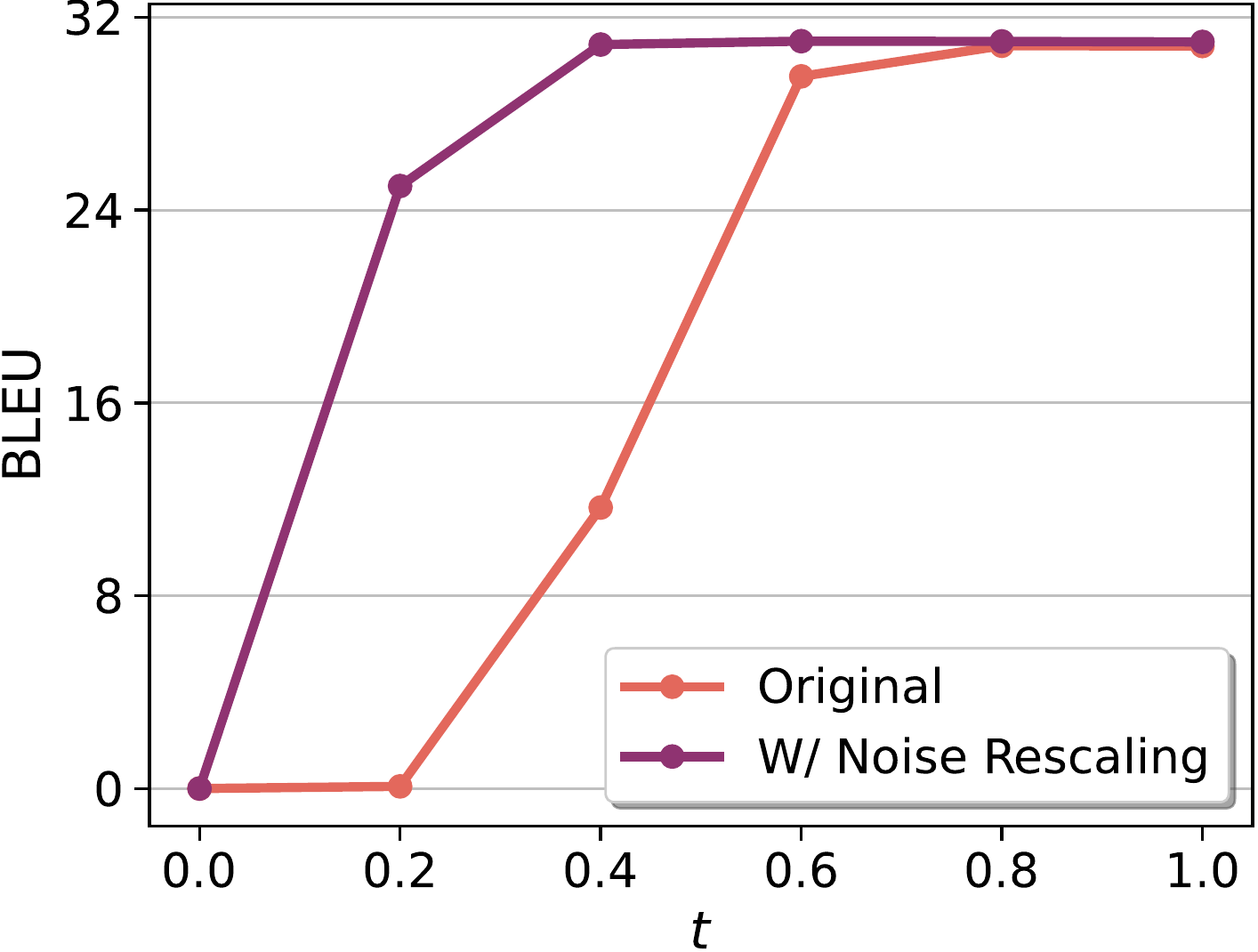}
        \caption{}
        \label{fig:bleu}
    \end{subfigure}%
    \hfill%
    \begin{subfigure}{0.3\textwidth}
        \includegraphics[width=\textwidth]{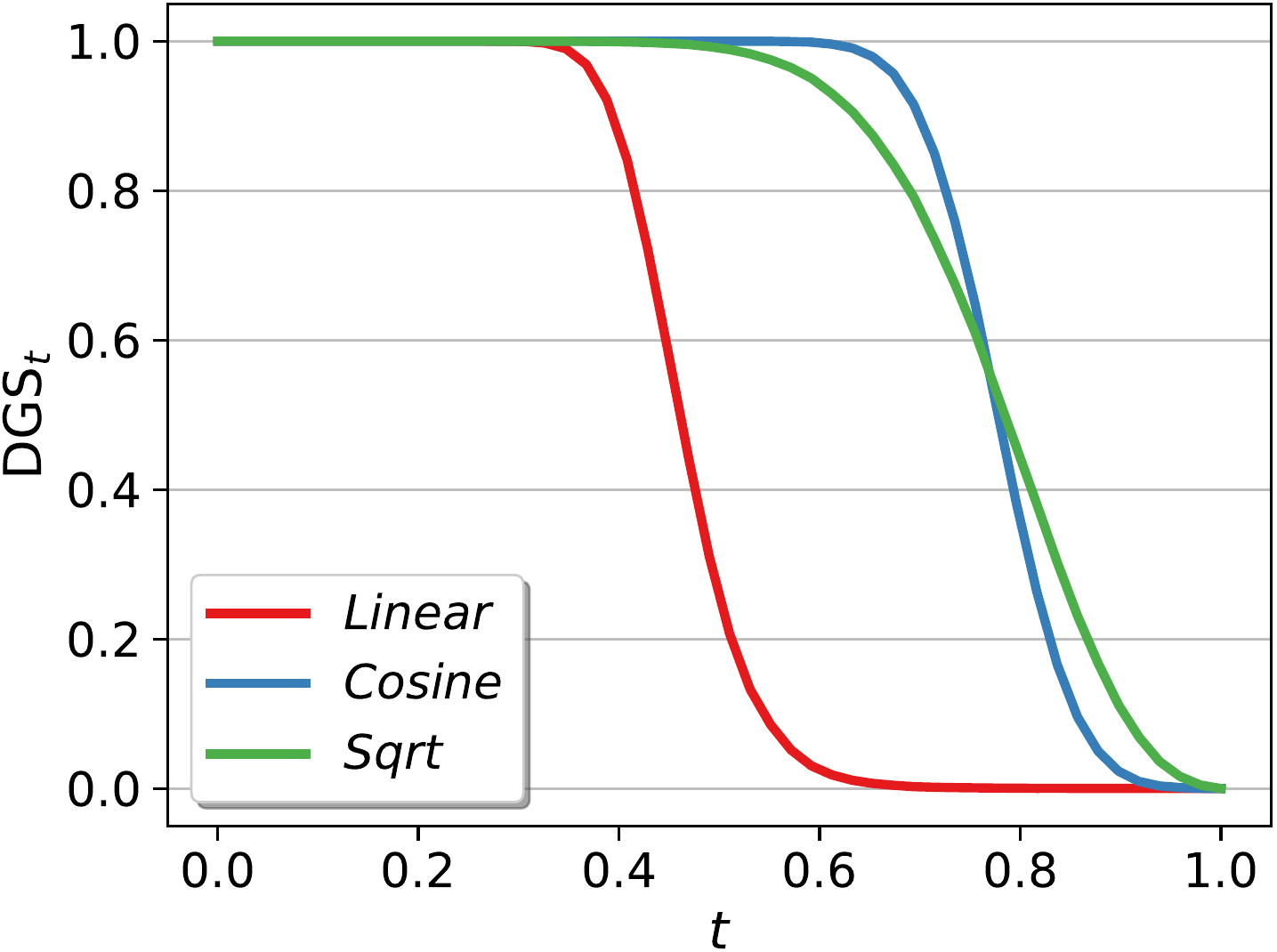}
        \caption{}
        \label{fig:dgs}
    \end{subfigure}%
    \hfill%
    \begin{subfigure}{0.3\textwidth}
        \includegraphics[width=\textwidth]{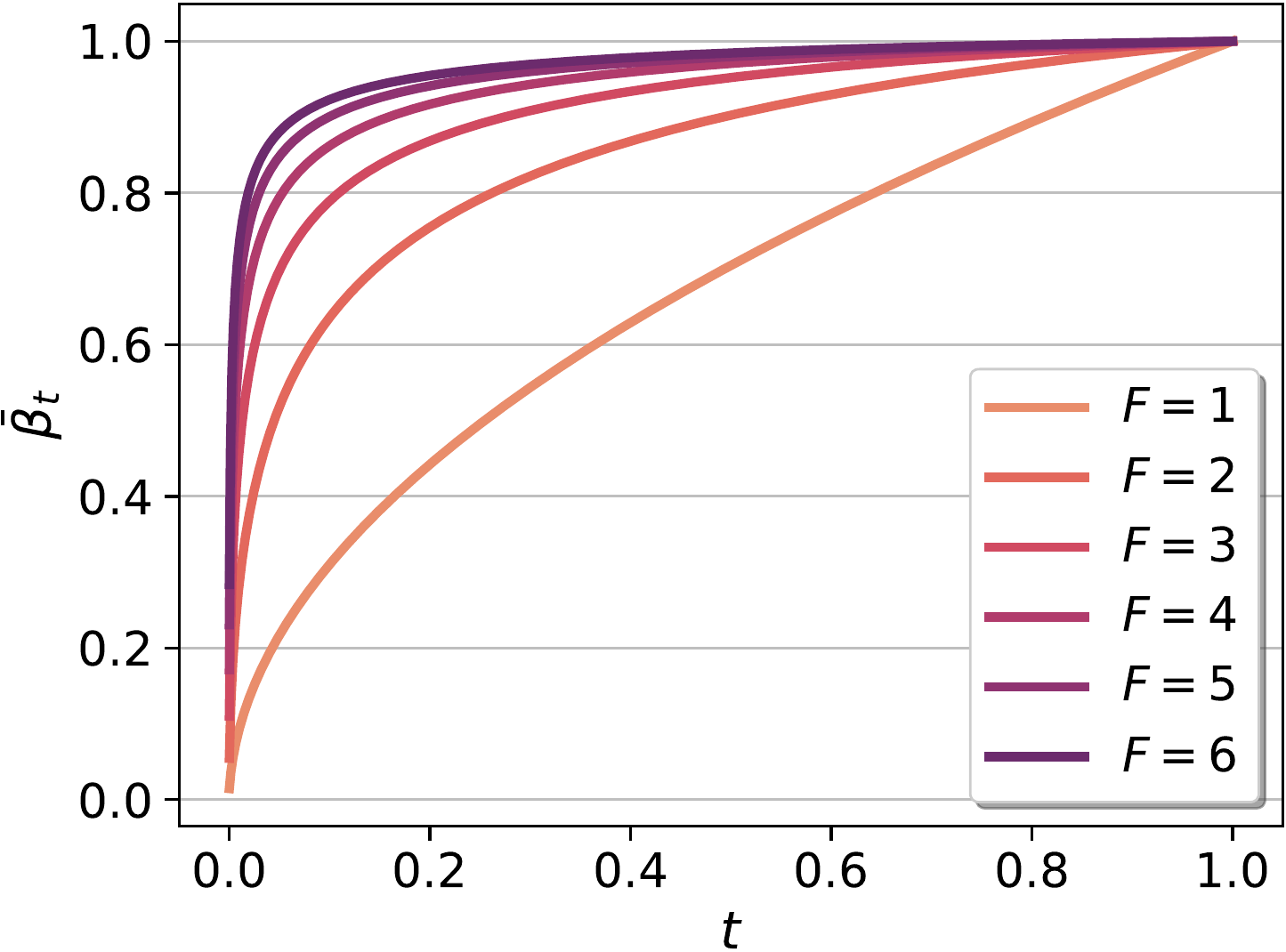}
        \caption{}
        \label{fig:schedule}
    \end{subfigure}
    \hspace*{\fill}%
    \caption{
        (a) BLEU score of models fed with pure Gaussian noise $\z_T$ on the IWSLT14 De-En dataset. The value of $t$ is normalized to $[0, 1]$.
        (b) $\dgs_t$ with different widely used schedules.
        (c) The \emph{sqrt} schedule rescaled with different values of the rescaling factor.
    }
\end{figure*}

\subsection{Degeneration of the Denoising Model}
\label{sec:degeneration}

\paragraph{Analysis of the Degeneration Problem}

The design of the noise schedule, which determines the amount of noise added to the data at each step, has significant influences on both forward and reverse processes.
Intuitively, denoising is a more challenging task for the model with higher levels of noise, and becomes easier when insufficient corruption is applied, where the model can generate the correct embeddings without depending on the condition and context. As a consequence, the model tends to degenerate to a trivial solution.
Here, we provide in-depth analyses of this problem. We start by defining the degenerated model, which discards the conditioning information and generates each embedding by choosing the nearest ones independently:
\begin{definition}
    \label{def:trivial}
    For a noised input $\z_t = [\zvec_{t, 1}, \zvec_{t, 2}, \cdots, \zvec_{t, n}]$, the \textbf{Degenerated Model} is defined as
    \begin{equation*}
        \fdg(\z_t; \xhid)
        = \left[\argmin_{\e_y \in \eset} \lours(\zvec_{t, i}, \e_y) \right]_i^n,
    \end{equation*}
    where
    \begin{equation*}
        \lours(\zvec_{t, i}, \e_y)
        = \lVert \zvec_{t, i} - \e_y \rVert^2
        - \log p_{\phi}(y | \zvec_{t, i}).
    \end{equation*}
\end{definition}
It can be proved that when insufficient noise is introduced during training, the denoising model tends to fall as the degenerated model defined above.
\begin{theorem}
    \label{thm:minimum}
    Given embeddings $\emat \sim \mathcal{N}_{d \times V} (\mathbf{0}, \sigma_e \mathbf{I})$, the probability of the degenerated model being a global minimum of the objective function $\lours$ for $\theta$ converges to $1$ as $\betabar \to 0$ and $d \to \infty$.
\end{theorem}
We leave the proof and illustrations of this theorem in \cref{sec:proof}.

This phenomenon could be verified quantitatively. To analyze the capacity of the denoising model at each noise level, we evaluate the BLEU score of $\zhat$ generated by the model at different timesteps.
To eliminate the impact of the noise schedule, we feed the model with $z_T$, \ie, the pure noise, rather than $z_t$.
As illustrated in \cref{fig:bleu}, the BLEU score drops dramatically at $t$s with low noise levels, and the range with low scores occupies nearly half of the axis. In other words, the model degenerates significantly and extensively, implying the noise levels brought by the schedule are far from being sufficient.

\paragraph{Noise Rescaling}

In most text generation tasks, a degenerated model is undesirable, as it fails to maintain contextual coherence and condition consistency, both of which are crucial for the task.
We propose a universal noise rescaling framework to alleviate the degeneration.
To achieve this, we start by defining the degree of degeneration to guide the rescaling. Intuitively, as supported by the proof of \cref{thm:minimum}, the tendency toward degeneration is highly related to the accuracy of the degenerated model. Thus we define a mathematical representation of the overall degree of degeneration:

\begin{definition}
    The \textbf{DeGeneration Score~(DGS)} of a specific noise schedule is
    \begin{equation*}
        \dgs (\alphabar, \betabar)
        = \frac1T \int_0^T \dgs_t (\alphabar_t, \betabar_t) \,\mathrm{d}t,
    \end{equation*}
    where $\dgs_t$ is the classification accuracy of the degenerated model at $t$
    \begin{equation*}
        \dgs_t = P(\fdg(\zvec_t) = \zvec_0),
    \end{equation*}
    $\alphabar$ and $\betabar$ are functions or discrete series representing the noise schedule, and $\zvec_t \sim \mathcal{N} \left( \sqrt{\alphabar_t} \zvec_0, \betabar_t \mathbf{I} \right)$. Note that we do not set the constraint $\alphabar_t + \betabar_t = 1$ here.
\end{definition}

\cref{fig:dgs} illustrates $\dgs_t$ of several widely utilized noise schedules, including \emph{linear}~\citep{ho2020denoising}, \emph{cosine}~\citep{nichol2021improved} and \emph{sqrt}~\citep{li2022diffusion}, with $\dgs$ listed in \cref{tab:dgs}.
From the table, we can notice that the schedule with a lower degeneration score yields a better BLEU score, which reflects the relationship between the degeneration score and the overall performance.

\begin{table}[t]
    \small
    \centering
    
    \begin{tabular}{lcc}
        \toprule
        \textbf{Schedules} & $\dgs$ & BLEU \\
        \midrule
        
        \emph{Linear} & $0.47$ & $32.21$ \\
        \emph{Cosine} & $0.77$ & $26.61$ \\
        \emph{Sqrt} & $0.77$ & $22.70$ \\
        \bottomrule
    \end{tabular}
    \caption{The degeneration score of several noise schedules proposed by prior works, and their performance on the IWSLT14 De-En dataset.}
    \label{tab:dgs}
\end{table}

Based on the degeneration score, we first specify a threshold $\dgsmax$ to the degeneration score, to impose a restriction on the noise added to the embeddings, under which we expect the model will not degenerate.
Then we introduce a factor to the noise schedule named the rescaling factor, amplifying the noise added to the embeddings to ensure that the noise schedule satisfies the restriction imposed by $\dgsmax$,
which can be written as
\begin{equation}
    \label{equ:rf}
    \left\{
        \begin{aligned}
            \alphabar_t' &= \alphabar_t  \\
            \betabar_t' &= F^2 \betabar_t 
        \end{aligned}
    \right.,
\end{equation}
where $\alphabar_t$ and $\betabar_t$ denote the original schedules, $\alphabar_t'$ and $\betabar_t'$ denote the schedule coefficients after rescaling, and $F$ is the rescaling factor.
Through experiments in \cref{sec:ablation}, we find this simple but effective adjustment brings significant improvement. In \cref{sec:search_rf}, we present an approach for searching $F$ given a specific $\dgsmax$, accompanied by a pre-computed function table to facilitate future research.
Alternatively, we can also derive a variance-preserving~\citep{song2020score} variant of the rescaling factor, which satisfies the constraint $\alphabar_t' + \betabar_t' = 1$ (details in \cref{sec:vprf}).
In \cref{fig:schedule} we show the shape of the \emph{sqrt} schedule rescaled by different values of $F$. As demonstrated in \cref{fig:bleu}, with our rescaling technique, the degeneration of the denoising model experiences a substantial alleviation in both degree and occupancy.

\subsection{Difformer}

Based on the analysis of the challenges encountered with embedding diffusion models, we introduce Difformer, a denoising diffusion model with Transformer, with the proposed techniques including the anchor loss and the noise rescaling technique.
An overview of the model is demonstrated in \cref{fig:overview}.

\paragraph{Length Prediction and 2D Parallel Decoding}
\label{sec:2d}

Unlike traditional autoregressive models where the sequence length is implicitly decided by the EOS token, diffusion models generate all tokens in a non-autoregressive manner, where the length should be modeled explicitly. Previous works~\citep{li2022diffusion,gong2022diffuseq,strudel2022selfconditioned} usually generate a sequence with the maximum length and cut off the content after the EOS token. In this paper, we utilize a more efficient way by explicitly predicting the target length with the encoder output~\citep{lee2018deterministic}, \ie, $p_{\theta}(n | \xhid)$, and a negative log-likelihood loss function is added to \cref{equ:ours} while training.

A unique benefit of this approach is that we can conduct 2D parallel decoding in inference. Firstly, we can consider top-$b_1$ lengths from the length predictor to generate candidates with different lengths. Secondly, for each length, we can also generate $b_2$ candidates by sampling different initial noises from the prior. The final prediction is selected from the total $b = b_1 \times b_2$ candidates that minimize the expected risk~\citep{kumar2004minimum} \wrt a metric such as BLEU or PPL.
We term two kinds of beams as length beam and noise beam respectively. We conduct a study on the impact of $b_1$ and $b_2$ in \cref{sec:beam}.
In summary, both beams introduce improvements, and $b_1$ influences more.

\paragraph{Acceleration in Inference}

Diffusion models are trained with thousands of forward steps, 
but it would be extremely time-consuming
to iterate all steps in inference. For Difformer, we pick a subset $\{ \tau_1, \tau_2, \cdots, \tau_K \}$ of size $K$ from the full diffusion trajectory $\{ 1, 2, \cdots, T \}$ for generation~\citep{song2020denoising, nichol2021improved}.
Correspondingly, the generated sample should be drawn from $q(\z_{\tau_{i - 1}} | \z_{\tau_i}, \zhat(\z_{\tau_i}, \xhid, \tau_i))$.
In consequence, the time complexity of generation is reduced from $\mathcal{O}(T)$ to $\mathcal{O}(K)$.

\begin{table*}[t]
    \small
    \centering
    \begin{tabular}{l *{5}{c}}
        \toprule
        \multirow{2}{*}{\textbf{Models}} & \multirow{2}{*}{$b$} &
        \textbf{WMT14 En-De} & \textbf{WMT16 En-Ro} & \textbf{IWSLT14 De-En} & \textbf{Gigaword} \\
        && \textbf{BLEU} & \textbf{BLEU} & \textbf{BLEU} & \textbf{ROUGE-$1$/$2$/L} \\
        \midrule
        
        Transformer & $1$ &
        $26.37$ & $32.76$ & $32.62$ & $36.78$/$17.79$/$34.10$ \\
        Transformer & $5$ &
        $27.37$ & $33.59$ & $33.91$ & $37.54$/$18.80$/$34.93$ \\
        
        CMLM & $1$ &
        $26.56$* & $32.75$* & $26.41$ & $34.41$/$15.61$/$32.17$ \\
        CMLM & $5$ &
        $27.03$* & $33.08$* & $31.76$ & $36.33$/$17.82$/$33.83$ \\
        \midrule
        
        DiffuSeq & $1$ &
        $13.73$ & $23.37$ & $27.03$ & $28.50$/$10.10$/$26.00$ \\
        DiffuSeq & $10$ &
        $15.37$ & $25.45$ & $28.78$ & $31.17$/$12.23$/$29.24$ \\
        SeqDiffuSeq & $1$ &
        $23.63^\dag$ & $23.98$ & $28.65$ & $30.28$/$11.72$/$28.40$ \\
        SeqDiffuSeq & $10$ &
        $24.24^\dag$ & $26.17$ & $30.03$ & $31.90$/$12.36$/$29.22$ \\

        DiNoiSer & $5$ &
        $26.08^\ddag$ & $32.57^\ddag$ & $32.23^\ddag$ & - \\
        DiNoiSer & $50$ &
        $26.29^\ddag$ & $32.59^\ddag$ & $32.48^\ddag$ & - \\
        \midrule
        
        \textbf{Difformer} & $1$ &
        $26.74^\Uparrow$ & $32.52^\Uparrow$ & $32.91^\Uparrow$ & $35.45$/$16.46$/$32.87^\Uparrow$ \\
        \textbf{Difformer} & $10$ &
        $27.70^\Uparrow$ & $33.18^\Uparrow$ & $\textbf{34.48}^\Uparrow$ & ${37.12}$/${18.25}$/${34.60}^\Uparrow$ \\
        \textbf{Difformer} & $20$ &
        $\textbf{27.74}$  & $\textbf{33.36}$ & $\textbf{34.48}$ & $\textbf{37.64}$/$\textbf{18.75}$/$\textbf{35.01}$ \\
        \bottomrule
    \end{tabular}
    \caption{
        The performance of the proposed Difformer and the baseline methods.
        *, $\dag$ and $\ddag$ indicate results reported by \citet{ghazvininejad2019mask}, \citet{yuan2023seqdiffuseq} and \citet{ye2023dinoiser} respectively.
        Other results are from our implementation.
        $\Uparrow$ indicates that Difformer outperforms all diffusion-based baselines with the same beam size $b$.
    }
    \label{tab:main_results}
\end{table*}

\begin{table*}[t]
    \small
    \centering
    \begin{tabular}{lc*{9}{C{2.5em}}}
        \toprule
        \multirow{2}{*}{\textbf{Models}} & \multirow{2}{*}{$b$} &
        \multicolumn{3}{c}{\textbf{QQP}} &
        \multicolumn{3}{c}{\textbf{Wiki-Auto}} &
        \multicolumn{3}{c}{\textbf{QT}} \\
        
        \cmidrule(lr){3-5} \cmidrule(lr){6-8} \cmidrule(lr){9-11} &&
        \textbf{B} & \textbf{R-L} & \textbf{BS} &
        \textbf{B} & \textbf{R-L} & \textbf{BS} &
        \textbf{B} & \textbf{R-L} & \textbf{BS} \\
        \midrule
        
        Transformer & $1$ &
        $29.65$ & $59.88$ & $84.28$ &
        $41.68$ & $58.15$ & $81.40$ &
        $16.83$ & $35.87$ & $63.97$ \\
        
        Transformer & $5$ &
        $30.83$ & $61.20$ & $85.29$ &
        $43.86$ & $58.48$ & $81.71$ &
        $16.45$ & $35.59$ & $63.91$ \\
        \midrule
        
        DiffuSeq & $10$ &
        $24.13$ & $58.80$ & $83.65$ &
        $36.22$ & $58.49$ & $81.26$ &
        $17.31$ & $\textbf{36.65}$ & $61.23$ \\
        
        SeqDiffuSeq & $1$ &
        $23.28$ & - & $82.91$ &
        $37.09$ & - & $82.11$ &
        $\textbf{17.20}$ & - & $61.35$ \\
        
        SeqDiffuSeq & $10$ &
        $24.34$ & - & $84.00$ &
        $37.12$ & - & $82.14$ &
        $17.46$ & - & $61.74$ \\

        DiNoiSer & $10$ &
        $26.07$ & - & - &
        $35.36$ & - & - &
        - & - & - \\

        DiNoiSer & $20$ &
        $25.42$ & - & - &
        $36.94$ & - & - &
        - & - & - \\
        \midrule
        
        \textbf{Difformer} & $1$ &
        $28.52^\Uparrow$ & $60.15$ & $83.80^\Uparrow$ &
        $40.37^\Uparrow$ & $59.56$ & $81.96$ &
        $16.03$ & $35.06$ & $61.05$ \\
        
        \textbf{Difformer} & $10$ &
        $30.43^\Uparrow$ & $\textbf{61.25}^\Uparrow$ & $\textbf{85.02}^\Uparrow$ &
        $40.77^\Uparrow$ & $59.86^\Uparrow$ & $82.21^\Uparrow$ &
        $16.66$ & $36.15$ & $63.29^\Uparrow$ \\
        
        \textbf{Difformer} & $20$ &
        $\textbf{30.52}^\Uparrow$ & $61.08$ & $\textbf{85.02}$ &
        $\textbf{40.84}^\Uparrow$ & $\textbf{59.88}$ & $\textbf{82.29}$ &
        $16.88$ & $36.28$ & $\textbf{63.32}$ \\
        \bottomrule
    \end{tabular}
    \caption{
        The performance of the proposed Difformer on the QQP, Wiki-Auto, and QT datasets.
        B, R-L, and BS stand for the BLEU, ROUGE-L, and BERTScore respectively.
        The results of DiffuSeq, SeqDiffuSeq, and DiNoiSe are from their paper. 
        $\Uparrow$ indicates that Difformer outperforms all diffusion-based baselines with the same beam size $b$.
    }
    \label{tab:other_results}
\end{table*}

\section{Experiments}

\subsection{Experimental Setup}

To evaluate the proposed Difformer model, we conduct experiments on five conditional text generation tasks including machine translation, text summarization, paraphrasing, text simplification, and question generation.

\paragraph{Datasets}

For machine translation, mainly following previous works~\citep{gu2018non, guo2019non, ghazvininejad2019mask}, three benchmark datasets WMT14 En-De~\citep{bojar2014findings}, WMT16 En-Ro~\citep{bojar2016findings} and IWSLT14 De-En~\citep{cettolo2014report} are inclued. For text summarization, experiments are conducted on Gigaword~\citep{graff2003english, rush2015neural}.
In addition, following previous non-autoregressive text generation works~\citep{gu2018non, gu2019levenshtein, ghazvininejad2019mask}, for machine translation and text summarization tasks, we adopt sequence-level knowledge distillation~\citep{kim2016sequence} on the original training set to alleviate the multi-modality problem.
For paraphrasing, text simplification, and question generation tasks, we mainly follow~\citep{gong2022diffuseq} to conduct experiments on Quora Question Pairs~(QQP)\footnote{\url{https://www.kaggle.com/c/quora-question-pairs}}, Wiki-Auto~\citep{jiang2020neural} and Quasar-T~(QT)~\citep{dhingra2017quasar} respectively.
The data split of the above datasets can be found in \cref{sec:settings}.

\paragraph{Metrics}

We report the tokenized BLEU and the SacreBLEU~\citep{post2018call} for machine translation tasks,
and the ROUGE~\citep{lin2004rouge} for summarization.
As for paraphrasing, text simplification, and question generation tasks, tokenized BLEU, ROUGE-L, and BERTScore~\citep{zhang2019bertscore} are utilized.

\paragraph{Baselines}

We mainly compare our method with recent embedding diffusion models, including DiffuSeq~\citep{gong2022diffuseq}, SeqDiffuSeq~\citep{yuan2023seqdiffuseq}, and DiNoiSer~\citep{ye2023dinoiser}, which extend Diffusion-LM~\citep{li2022diffusion} to the sequence-to-sequence scenario.
We further compare to a recent score-based model CDCD~\citep{dieleman2022continuous}.
CMLM~\citep{ghazvininejad2019mask} is also included, a non-autoregressive model with iterative decoding, which can be considered as a discrete diffusion model~\citep{austin2021structured}.
In addition, we report the performance of Transformer as the autoregressive baseline.

\paragraph{Implementation Details}

We set diffusion step $T = 2000$, embedding dimension $d = 128$, the threshold of the degeneration score $\dgsmax = 0.15$, and use the \emph{sqrt} noise schedule. Following previous works~\citep{li2022diffusion, strudel2022selfconditioned}, we also utilize the self-conditioning~\citep{chen2022analog} which is shown effective in improving the final performance.
More details of experiment settings can be found in \cref{sec:settings}.

\subsection{Results}

The main results are listed in \cref{tab:main_results,tab:other_results}.
With a little abuse of notation, we use $b$ to represent the size of beam search for the Transformer baseline, as well as the size of parallel decoding~(\ie, $b=b_1 \times b_2$).
As can be observed from experimental results, the proposed Difformer outperforms both the diffusion-based and iteration-based non-autoregressive baselines on most of the datasets with different choices of $b$, and even performs comparably with the autoregressive Transformer model. 
Specifically, the significant improvements of Difformer over the diffusion baselines confirm the challenges that occur to embedding diffusion models for text generation tasks and the effectiveness of the proposed solutions. Compared with CMLM, an iteration-based non-autoregressive baseline, Difformer outperforms on various datasets consistently. 
Moreover, benefiting from the stochastic nature of diffusion models, Difformer is able to conduct 2D parallel decoding over the length and noise beam at the same time, increasing its flexibility and potential to obtain better results.
We further compare with baselines on the raw and distilled WMT14 En-De training set with SacreBLEU\footnote{The signature is \texttt{nrefs:1|\-case:mixed|\-eff:no|\-tok:13a|\-smooth:exp|\-version:2.2.0}} as the metric. As shown in \cref{tab:raw_sacrebleu} and \cref{tab:distilled_sacrebleu}, the proposed Difformer achieves better results with the same number of $b$. Due to the page limit, we leave results of more metrics and baselines in \cref{sec:metrics,sec:baselines}, which also validate the superiority of Difformer.

\begin{table}[t]
    \small
    \centering
    
    \begin{tabular}{lcc}
        \toprule
        \textbf{Models} & $b$ & \textbf{SacreBLEU} \\
        \midrule
        
        CDCD & $1$ & $19.30$ \\
        CDCD & $10$ & $19.70$ \\
        SeqDiffuSeq & $1$ & $19.16$ \\
        SeqDiffuSeq & $10$ & $19.76$ \\
        DiNoiSer & $5$ & $24.25$ \\
        DiNoiSer & $50$ & $24.62$ \\
        \midrule
        
        \textbf{Difformer} & $1$ & $22.80$ \\
        \textbf{Difformer} & $10$ & $24.10$ \\
        \textbf{Difformer} & $50$ & $\textbf{24.90}$ \\
        \bottomrule
    \end{tabular}
    \caption{
        SacreBLEU scores on the raw WMT14 En-De dataset.
        The results of the baselines are as reported in their paper.
    }
    \label{tab:raw_sacrebleu}
\end{table}

\begin{table}[t]
    \small
    \centering
    
    \begin{tabular}{lcc}
        \toprule
        \textbf{Models} & $b$ & \textbf{SacreBLEU} \\
        \midrule
        
        DiNoiSer & $5$ & $25.70$ \\
        DiNoiSer & $50$ & $25.90$ \\
        \midrule
        
        \textbf{Difformer} & $10$ & $\textbf{26.20}$ \\
        \bottomrule
    \end{tabular}
    \caption{
        SacreBLEU scores on the distilled WMT14 En-De dataset.
        The results of DiNoiSer are as reported in their paper.
    }
    \label{tab:distilled_sacrebleu}
\end{table}

\subsection{Analyses}

\paragraph{Ablation Study}
\label{sec:ablation}

We study the effects of the proposed components, which are listed in \cref{tab:ablation}.
Firstly, while previous embedding diffusion works usually utilize the rounding loss function, we find it does not provide satisfactory results, which echoes our findings in \cref{sec:collapse}. By replacing $\lround$ with $\lanchor$, the problem is largely alleviated with significant performance improvements.
The enhancement from noise rescaling also reinforces our findings in \cref{sec:degeneration} that the model suffers from a degeneration problem.
Besides, the integration of the anchor loss and noise rescaling yields the best performance.

\begin{table}[t]
    \small
    \centering

    \begin{tabular}{
        C{3em}
        C{5em}
        C{3em}
    }
        \toprule
        \textbf{Anchor Loss} & \textbf{Noise Rescaling} & \textbf{BLEU} \\
        \midrule
        
        && $16.96$ \\
        \checkmark && $22.70$ \\
        & \checkmark & $27.89$ \\
        \checkmark & \checkmark & $\textbf{34.48}$ \\
        \bottomrule
    \end{tabular}
    \caption{The ablation study on the proposed components. Results are conducted on the IWSLT14 De-En dataset with $b=10$.}
    \label{tab:ablation}
\end{table}

\paragraph{Inference Speed}

Continuous diffusion models usually rely on hundreds or thousands of reverse steps in inference to guarantee the quality of the generated samples~\citep{li2022diffusion, gong2022diffuseq, strudel2022selfconditioned}.
In contrast, we find that Difformer is able to achieve considerably good performance with much fewer reverse steps.
In \cref{tab:speed}, we evaluate the BLEU score and inference speed by varying the number of reverse steps.
The conclusions are two-fold. Firstly, Difformer performs robustly \wrt $K$, especially compared with diffusion-based baselines. We attribute this advantage to the anchor loss, as it facilitates learning a well-distributed embedding space, and connects $\zhat$ with solid ground truth labels, which reduce the obscurity of predictions.
Correspondingly, the inference speed of Difformer outperforms the autoregressive model Transformer by $6$ times and the iterative non-autoregressive model CMLM by $3$ times when $K$ is small, showing the potential of deploying Difformer to online systems.

\begin{table}[t]
    \small
    \centering
    \begin{tabular}{
        lccc
    }
        \toprule
        \textbf{Models} & $K$ & \textbf{Speed} & \textbf{BLEU} \\
        \midrule

        Transformer & $n$ & $6.05$ & $33.91$ \\
        CMLM & $10$ & $11.80$ & $31.76$ \\
        \midrule

        DiffuSeq & $2000$ & $0.06$ & $28.78$ \\
        DiffuSeq & $1000$ & $0.12$ & $23.91$ \\
        DiffuSeq & $500$ & $0.23$ & $0.96$ \\
        SeqDiffuSeq & $2000$ & $0.05$ & $30.03$ \\
        \midrule
        
        Difformer & $2000$ & $0.03$ & $34.09$ \\
        Difformer & $20$ & $6.30$ & $34.19$ \\
        Difformer & $10$ & $11.40$ & $34.13$ \\
        Difformer & $1$ & $39.51$ & $30.14$ \\
        \midrule
        
    \end{tabular}
    \caption{The inference speed and corresponding performance of the proposed Difformer and the baselines. The speed is represented as sentences per second. Results are conducted on the IWSLT14 De-En with $b = 10$ and batch size $ = 1$, without early stopping~(see~\cref{sec:dynamics}).}
    \label{tab:speed}
\end{table}

\paragraph{Diversity}

\begin{figure}[t]
    \centering
    \includegraphics[width=0.8\columnwidth]{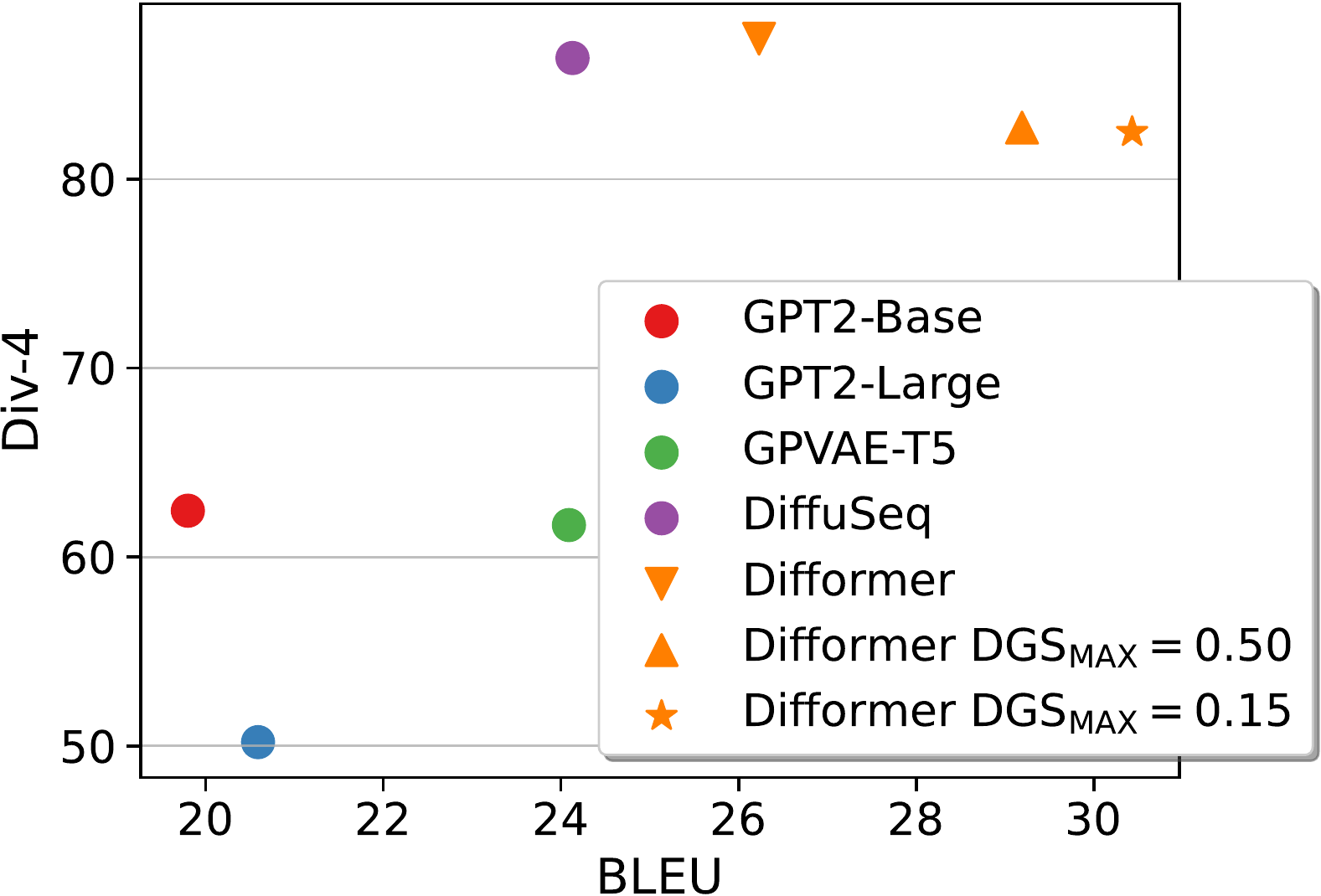}
    \caption{
        Comparison of generation quality and diversity of baselines and Difformer on the QQP dataset. Results of baselines are from \citep{gong2022diffuseq}.
    }
    \label{fig:bleu_div}
\end{figure}

To conduct a comprehensive evaluation of Difformer, we incorporate 4-gram diversity~(div-4)~\citep{deshpande2019fast} as the diversity metric on the QQP dataset, and expand our comparison with LLMs, including GPT-2~\citep{radford2019language} and GPVAE-T5~\citep{du2022diverse}. According to~\cref{fig:bleu_div}, we observe that the adjustment of $\dgsmax$ introduces a trade-off between generation quality and diversity. Specifically, the model trained with a smaller $\dgsmax$ manifests reduced diversity but improved quality. We assume that this is attributed to the stronger restrictions imposed by this $\dgsmax$ leading to a higher determinacy. Notably, under the condition 
$\dgsmax = 0.15$ and without noising rescaling, Difformer demonstrates superior performance over the baselines including LLMs in terms of quality and diversity respectively.

\paragraph{Generation Quality Dynamics}
\label{sec:dynamics}

\begin{figure}[t]
    \centering
    \includegraphics[width=0.8\columnwidth]{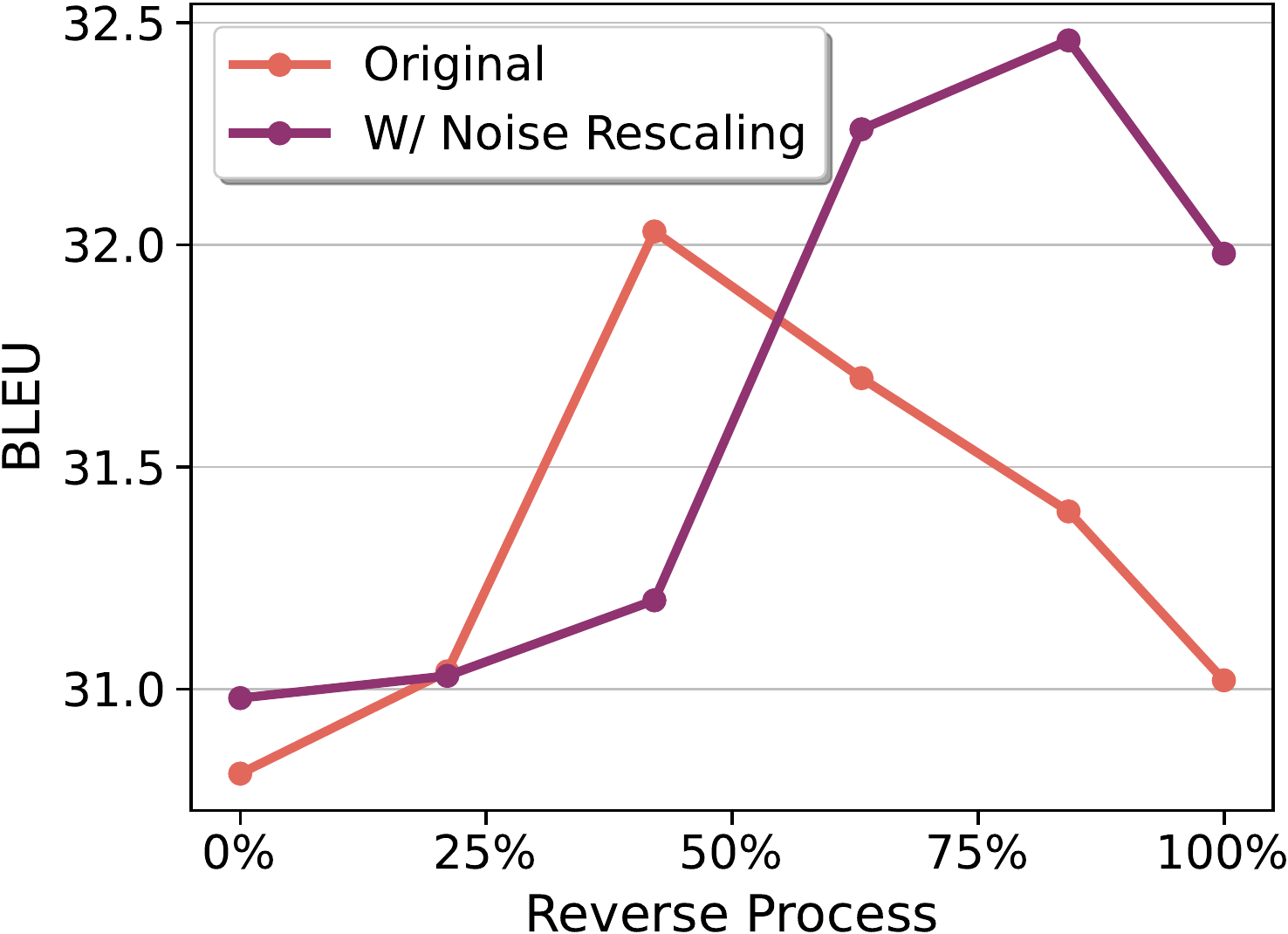}
    \caption{The intermediate BLEU score of $\zhat$ within a decoding process.}
    \label{fig:dynamics}
\end{figure}

To investigate the dynamics of generation quality during the reverse process, we extract $\zhat$ at intermediate reverse steps and evaluate their BLEU scores. The results are illustrated in \cref{fig:dynamics}.
For the original noise, in the first third of the reverse process, the generation quality continuously increases as expected. However, there is a noticeable performance decline in the latter part, where the model degenerates gradually.
When noise rescaling is applied, a notable improvement can be observed, manifesting that the degeneration problem is alleviated as discussed in \cref{sec:degeneration}.
This observation also motivates us to propose an early stopping technique, terminating the decoding process at a proper intermediate step to retain a high-quality output.

\paragraph{Threshold of the Degeneration Score}

We further analyze the influence of different values of $\dgsmax$ in \cref{tab:rescaling}. From the table, we can notice that, as $\dgsmax$ decreases, the BLEU score increases at first, and declines afterward. The former is reasonable since a small $\dgsmax$ mitigates the degeneration problem.
However, if $\dgsmax$ is too small, a large rescaling factor causes the noise schedule to become a constant, as illustrated in \cref{fig:schedule}. Therefore, the multistep denoising process degenerates into a one-step process.
In conclusion, a moderate $\dgsmax$ is sufficient to maintain a balance between the degeneration of the model and the denoising process. More results with schedules other than \textit{sqrt} can be found in \cref{sec:dgsmax_schedules}.

\begin{table}[t]
    \small
    \centering

    \begin{tabular}{ccc}
        \toprule
        \textbf{$\dgsmax$} & \textbf{Rescaling Factor} & \textbf{BLEU} \\
        \midrule
        
        $0.05$ & $7.0$ & $33.88$ \\
        $0.15$ & $4.0$ & $34.48$ \\
        $0.50$ & $2.0$ & $32.54$ \\
        $0.77$ & $1.0$ & $22.70$ \\
        \bottomrule
    \end{tabular}
    \caption{The performance with varying $\dgsmax$. The last row represents the model without noise rescaling. Results are obtained on the IWSLT14 De-En dataset with $b = 10$.}
    \label{tab:rescaling}
\end{table}

\section{Conclusion}

In this paper, we conduct a thorough study of the challenges when optimizing an embedding diffusion model on discrete textual data, and propose the corresponding solutions. Firstly, to tackle the challenge of embedding collapse and instability in training caused by the dynamic nature of embeddings,
we introduce an anchor loss to regularize the embeddings and stabilize the training simultaneously. Secondly, we derive a novel noise rescaling framework based on theoretical analysis, which notably alleviates the degeneration of the denoising model resulting from inadequate noise.
Finally, integrated with the aforementioned techniques, we present Difformer, a denoising diffusion model based on Transformer. Difformer demonstrates superior performance on various benchmark text generation tasks, outperforming prior diffusion-based models as well as iterative non-autoregressive models.

\section*{Limitations}

The improvement brought by the proposed techniques is promising. However, embedding diffusion models converge relatively slowly in training. For instance, compared with CMLM, Difformer requires around double the training time to reach convergence, although it is more efficient than prior diffusion models. Moreover, due to the cost of the search for the rescaling factor, it is performed offline and the factor is static during training.

\section*{Acknowledgements}

This research was supported by the National Natural Science Foundation of China (Grant No. 62276245).

\bibliography{custom}

\appendix

\section{Proof of \cref{thm:minimum}}
\label{sec:proof}

Following we give the proof of \cref{thm:minimum}. We start by proving two lemmas.
The first one gives the limit of the loss objective if provided insufficient small noise in a high-dimensional space.
And the second one proves that the accuracy of the degenerated model converges to $1$ on the same condition. Finally, \cref{thm:minimum} can be conducted.

\begin{lemma}
    \label{lem:loss}
    Assume the embeddings $\emat \sim \mathcal{N}_{d \times V} (\mathbf{0}, \sigma_e \mathbf{I})$, then
    \begin{equation*}
        \forall \varepsilon > 0,
        \lim_{\limcond}
        P(\lours(\zvec_t, \zvec_0) < \varepsilon) = 1,
    \end{equation*}
    which can be rewritten as 
    \begin{equation*}
        \lours(\zvec_t, \zvec_0) \convergesto 0.
    \end{equation*}

    Also, $\forall \e_i \in \eset$ and $\e_i \ne \zvec_0$,
    \begin{equation*}
        \lours(\zvec_t, \e_i) \convergesto 2\sigma_e^2.
    \end{equation*}
\end{lemma}

\begin{proof}
    \begin{align*}
        \small
        \lours(\zvec_t, \zvec_0) &= 
        \lVert \zvec_t - \zvec_0 \rVert^2
        - \log p_\phi(y_{\zvec_0} | \zvec_t) \\
        &= \lVert \zvec_t \rVert^2 + \lVert \zvec_0 \rVert^2 - \frac{2}{d}\zvec_t \cdot \zvec_0 \\
        &\quad - \log \frac{\exp(\zvec_t \cdot \zvec_0)}{\sum_{i = 1}^V \exp(\zvec_t \cdot \e_i)} \\
        &= \lVert \zvec_t \rVert^2 + \lVert \zvec_0 \rVert^2 - \frac{2}{d} \zvec_t \cdot \zvec_0 \\
        &\quad + \log \Bigg(
            1 \\
        &\quad + \sum_{i \ne y_{\zvec_0}}^V
            \exp(\zvec_t \cdot \e_i- \zvec_t \cdot \zvec_0)
        \Bigg),
    \end{align*}
    where $\lVert \cdot \rVert^2$ represents mean square error~(MSE), and $y_{\zvec_0}$ is the token index of $\zvec_0$.
    Since $\zvec_t \sim \mathcal{N} \left(\sqrt{1 - \betabar_t} \zvec_0, \betabar_t \mathbf{I}\right)$, which can be written as $\zvec_t = \sqrt{1 - \betabar_t} \zvec_0 + \sqrt{\betabar_t} \ep$, where $\ep \sim \mathcal{N}(\mathbf{0}, \mathbf{I})$, we get
    \begin{align}
        \label{equ:lt}
        \small
        \lours(\zvec_t, \zvec_0) &=
        \left( 2 - \betabar - 2\sqrt{1 - \betabar} \right) \lVert \zvec_0 \rVert^2
        + \betabar \lVert \ep \rVert^2 \nonumber \\
        &\quad + 2 \left( \sqrt{\betabar - \betabar^2} - \sqrt{\betabar} \right) \frac{\zvec_0 \cdot \ep}{d} \nonumber \\
        &\quad + \log \Bigg( 1 \\
        &\quad + \sum_{i \ne y_{\zvec_0}}^V
            \exp (g(\zvec_0, \ep, \e_i, \zvec_0) \cdot d)
        \Bigg),
    \end{align}
    where
    \begin{align}
        \small
        \label{equ:g}
        g(\boldsymbol{a}, \boldsymbol{b}, \boldsymbol{c}, \boldsymbol{d}) &=
        \sqrt{1 - \betabar} \frac{\boldsymbol{a} \cdot \boldsymbol{c}}{d}
        + \sqrt{\betabar} \frac{\boldsymbol{b} \cdot \boldsymbol{c}}{d} \nonumber \\
        &\quad - \sqrt{1 - \betabar} \frac{\boldsymbol{a} \cdot \boldsymbol{d}}{d}
        - \sqrt{\betabar} \frac{\boldsymbol{b} \cdot \boldsymbol{d}}{d}.
    \end{align}

    According to Khinchin's law,
    \begin{equation*}
        \lVert \zvec_0 \rVert^2 =
        \frac 1 d \sum_{i = 1}^d z_{0, i}^2
        \convergesto
        \mathbb{E} (z_{0, i}^2) = \sigma_e^2.
    \end{equation*}
    Similarly, the rest terms can be calculated. Based on the continuous mapping theorem, we can substitute these values into \cref{equ:g,equ:lt} as
    \begin{align*}
        \small
        g(\zvec_0, \ep, \e_i, \zvec_0)
        &\convergesto
        1 \cdot 0
        + 0 \cdot 0
        - 1 \cdot \sigma_e^2
        - 0 \cdot 0 \\
        &= -\sigma_e^2, \\
        \lours(\zvec_t, \zvec_0)
        &\convergesto
        0 \cdot \sigma_e^2 + 0 \cdot 1 + 0 \cdot 0 \\
        &\quad + \log (1 \\
        &\quad + (V - 1) \exp (-\sigma_e^2 \cdot \infty)) \\
        &= 0.
    \end{align*}

    Similarly,
    \begin{align*}
        \small
        \lours(\zvec_t, \e_i) &=
        \left( 1 - \betabar \right) \lVert \zvec_0 \rVert^2
        + \betabar \lVert \ep \rVert^2
        + \lVert \e_i \rVert^2 \\
        &\quad
        + 2 \sqrt{\betabar - \betabar^2} \frac{\zvec_0 \cdot \ep}{d} \\
        &\quad - 2 \sqrt{1 - \betabar} \frac{\zvec_0 \cdot \e_i}{d} \\
        &\quad - 2 \sqrt{\betabar} \frac{\ep \cdot \e_i}{d} \\
        &\quad
        + \log \Bigg( 1 \\
        &\quad + \sum_{j \ne i}^V
            \exp (g(\zvec_0, \ep, \e_j, \e_i) \cdot d)
        \Bigg) \\
        &\convergesto 2\sigma_e^2.
    \end{align*}
\end{proof}

\begin{lemma}
    \label{lem:degenerate}
    Assume the embeddings $\emat \sim \mathcal{N}_{d \times V} (\mathbf{0}, \sigma_e \mathbf{I})$, then for a single noised embedding $\zvec_t$, the output of the degenerated model
    \begin{equation*}
        \fdg(\zvec_t) \convergesto \zvec_0.
    \end{equation*}
\end{lemma}

\begin{proof}
    According to \cref{lem:loss}, $\forall \e_i \in \eset$ and $\e_i \ne \zvec_0$, we know
    \begin{equation*}
        \lours(\zvec_t, \zvec_0) \convergesto 0,
    \end{equation*}
    and,
    \begin{equation*}
        \lours(\zvec_t, \e_i) \convergesto 2\sigma_e^2.
    \end{equation*}

    Since $0 < 2\sigma_e^2$, according to the limiting inequality, 
    \begin{equation*}
        \lim_{\limcond}
        P(\lours(\zvec_t, \zvec_0) < \lours(\zvec_t, \e_i)) = 1,
    \end{equation*}
    which means
    \begin{equation*}
        \lim_{\limcond}
        P \left( \zvec_0 = \argmin_{\e \in \eset} \lours(\zvec_t, \e_i) \right) = 1.
    \end{equation*}

    Hence,
    \begin{equation*}
        \fdg(\zvec_t)
        = \argmin_{\e \in \eset} \lours(\zvec_t, \e_i)
        \convergesto \zvec_0.
    \end{equation*}
\end{proof}

\begin{proof}[Proof of \cref{thm:minimum}]
    From \cref{lem:degenerate},
    \begin{equation*}
        \fdg(\zvec_t) \convergesto \zvec_0.
    \end{equation*}
    Following the proof of \cref{lem:loss}, it is obvious that
    \begin{equation*}
        \lim_{\zvec \to \zvec_0} \lours(\zvec, \zvec_0)
        = \lours(\zvec_0, \zvec_0) \convergesto 0.
    \end{equation*}
    According to the law of the limit of compositions, 
    \begin{equation}
        \label{equ:loss0}
        \lours(\fdg(\zvec_t), \zvec_0) \convergesto 0.
    \end{equation}

    $\forall \theta'$, if
    \begin{equation*}
        \lours(f_{\theta'}(\zvec_t, t), \zvec_0)
        \convergesto
        \loss' > 0,
    \end{equation*}
    from limiting inequality,
    \begin{equation*}
        \lim_{\limcond}
        P(
            \lours(\fdg(\zvec_t), \zvec_0) <
            \lours(f_{\theta'}(\zvec_t, t), \zvec_0)
        ) = 1,
    \end{equation*}
    which means that the probability of $\fdg$ being a global minimum of $\lours$ for $\theta$ converges to $1$ as $\betabar \to 0$ and $d \to \infty$.

    Otherwise, if
    \begin{equation*}
        \lours(f_{\theta'}(\zvec_t, t), \zvec_0) \convergesto 0,
    \end{equation*}
    $f_{\theta'}$ satisfies the above conclusion that $\theta'$ is also a global minimum.

    Finally, if $\lours(f_{\theta'}(\zvec_t, t), \zvec_0)$ diverges, then $f_{\theta'}(\zvec_t, t)$ diverges, indicating $f_{\theta'}$ is an unstable model, or converges to infinity.
\end{proof}

\begin{figure}[t]
    \centering
    \includegraphics[width=0.8\columnwidth]{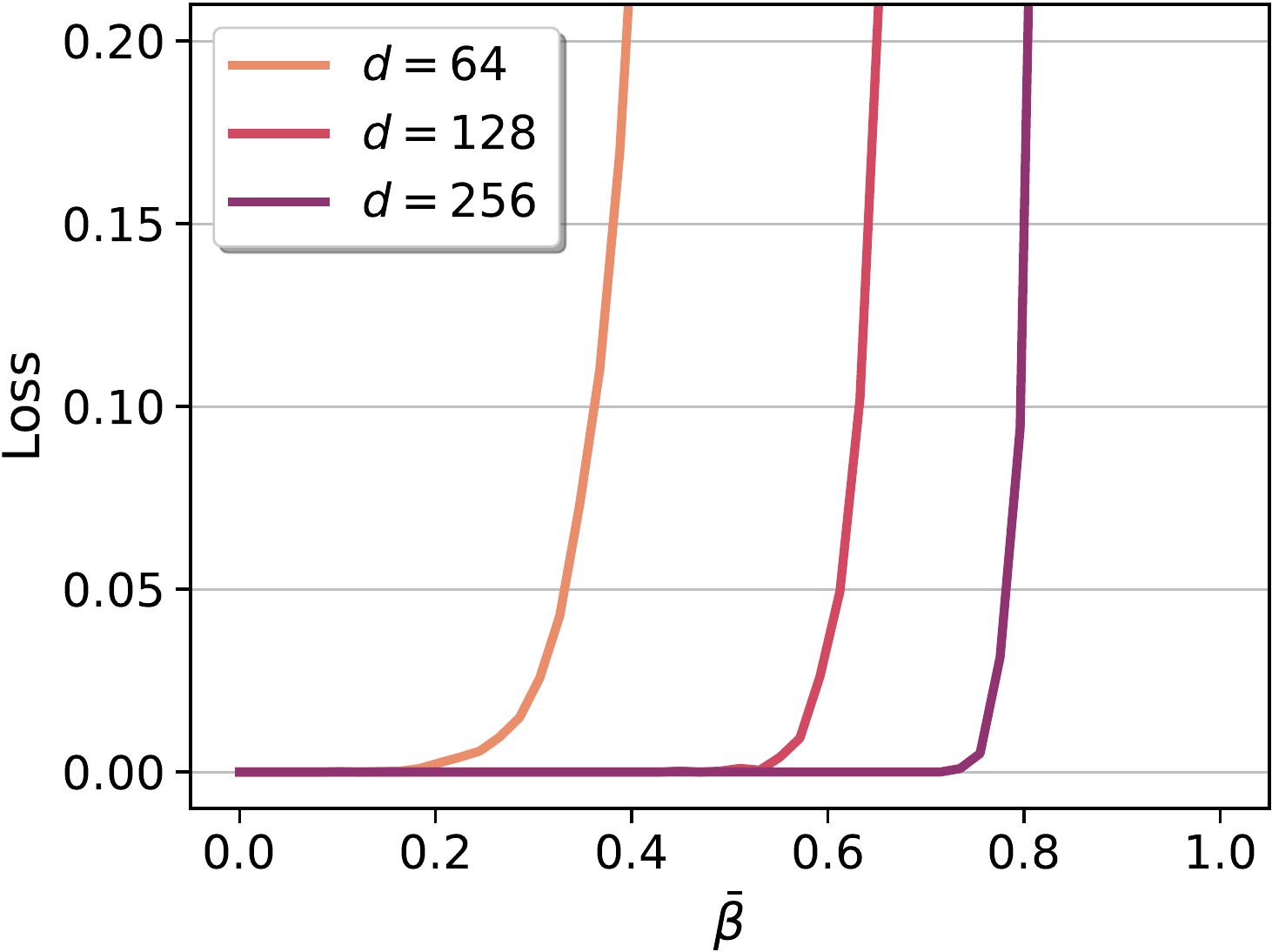}
    \caption{
        The loss of the degenerated model $\fdg$ with varying levels of noise and embedding dimensions.
    }
    \label{fig:deg_loss}
\end{figure}

Notably, though the condition $\betabar \to 0$ and $d \to \infty$ seem too strong, our empirical results exhibit a fast coverage speed of \cref{equ:loss0} in practice. \cref{fig:deg_loss} showcases the loss of the degenerated model, \ie, $\lours(\fdg(\zvec_t), \zvec_0)$ with varying levels of noise and embedding dimensions. According to the figure, even under a dimension of $64$, a $\betabar$ of $0.15$ is sufficiently small to ensure the loss converges to $0$.
Under higher dimensions, like $128$, we can observe that approximately $50\%$ of $\betabar$ obtains zero loss, which significantly emphasizes the tendency of degeneration. A similar tendency is also revealed by the model capacity decreasing in \cref{fig:bleu}.

\section{Detailed Derivation of the Objective}
\label{sec:derivation}

For a data sample $\z_0$, given a series of latent variables $\z_1, \cdots, \z_T$ which form a Markov chain, we start with the definition of the forward process:
\begin{equation*}
    q(\z_t | \z_{t - 1}) = \mathcal{N} \left(
        \z_t;
        \sqrt{\alpha_t} \z_{t - 1},
        \beta_t \mathbf{I}
    \right),
\end{equation*}
where $\alpha_t$ and $\beta_t$ represent the noise schedule, and $\alpha_t + \beta_t = 1$, $\z_T \sim \mathcal{N}(\mathbf{0}, \mathbf{I})$. Then the distribution of the latent variable at any timestep can be determined by
\begin{equation*}
    q(\z_t | \z_0) = \mathcal{N} \left(
        \z_t;
        \sqrt{\alphabar_t} \z_0,
        \betabar_t \mathbf{I}
    \right),
\end{equation*}
where $\alphabar_t := \prod_{i = 0}^t \alpha_i$ and $\betabar_t := 1 - \alphabar_t$.

The objective of diffusion models is to generate a denoising series to fit the reversion of the forward process, called the reverse process. The reverse process also forms a Markov chain, and is fixed to the learned Gaussian transitions
\begin{equation*}
    p_\theta(\z_{t - 1} | \z_t) = \mathcal{N} \left(
        \z_{t - 1};
        \boldsymbol{\mu}_{\theta}(\z_t, t),
        \mathbf{\Sigma}_\theta(\z_t, t)
    \right),
\end{equation*}
where $\boldsymbol{\mu}_\theta(\cdot)$ and $\mathbf{\Sigma}_\theta(\cdot)$ are learnable variables. The covariance is set to be a constant as $\mathbf{\Sigma}_\theta(\z_t, t) = \sigma_t^2 \mathbf{I}$ following \citet{ho2020denoising}.

The variational lower-bound can be derived from the negative log-likelihood as:
\begin{equation*}
    \small
    \mathbb{E} [-\log p_\theta(\z_0)]
    \leq \mathbb{E}_q \left[
        -\log \frac{p_\theta(\z_{0:T})} {q(\z_{1:T} | \z_0)}
    \right] = \lvlb.
\end{equation*}
Then according to the Markov property,
\begin{align*}
    \small
    \lvlb &=
    \mathbb{E}_q \left[
        -\log
        \frac{p(\z_T) \prod_{t = 1}^T p_\theta(\z_{t - 1} | \z_t)}
        {\prod_{t = 1}^T q(\z_t | \z_{t - 1})}
    \right] \\
    &= \mathbb{E}_q \left[
        -\sum_{t \ge 1} \log
        \frac{p_\theta(\z_{t - 1} | \z_t)}
        {q(\z_t | \z_{t - 1})}
    \right] + C_1 \\
    &= \mathbb{E}_q \Biggl[
        -\sum_{t > 1} \log
        \frac{p_\theta(\z_{t - 1} | \z_t)}
        {q(\z_{t - 1} | \z_t, \z_0)} \\
        &\quad -\log p_\theta(\z_0 | \z_1)
    \Biggr] + C_2 \\
    &= \mathbb{E}_q \Biggl[
        \sum_{t > 1} \mathbb{KL} [
            q(\z_{t - 1} | \z_t, \z_0) \Vert
            p_\theta(\z_{t - 1} | \z_t)
        ] \\
        &\quad -\log p_\theta(\z_0 | \z_1)
    \Biggr] + C_2,
\end{align*}
where $\mathbb{KL}[\cdot \Vert \cdot]$ denotes the KL divergence. 
Here, $q(\z_{t - 1} | \z_t, \z_0)$ has the closed form
\begin{equation*}
    q(\z_{t - 1} | \z_t, \z_0) = \mathcal{N} \left(
        \z_{t - 1};
        \tilde{\boldsymbol{\mu}}_t(\z_t, \z_0),
        \tilde{\beta}_t \mathbf{I}
    \right),
\end{equation*}
and
\begin{equation*}
    \left\{
        \begin{aligned}
            &\tilde{\boldsymbol{\mu}}_t(\z_t, \z_0) =
            \xi_t \z_0 + \lambda_t \z_t \\
            &\xi_t =
            \frac{\sqrt{\alphabar_{t - 1}} \beta_t}
            {1 - \alphabar_t} \\
            &\lambda_t =
            \frac{\sqrt{\alpha_t} (1 - \alphabar_{t - 1})}
            {1 - \alphabar_t}
        \end{aligned}
    \right..
\end{equation*}

Let
\begin{equation*}
    \loss_{t - 1} = \mathbb{KL} [
        q(\z_{t - 1} | \z_t, \z_0) \Vert
        p_\theta(\z_{t - 1} | \z_t)
    ],
\end{equation*}
using the formula for the KL divergence of Gaussian distributions, we can derive
\begin{equation*}
    \loss_{t - 1} =
    \frac{1}{2 \sigma_t^2} 
    \lVert
        \tilde{\boldsymbol{\mu}}_t(\z_t, \z_0)
        - \boldsymbol{\mu}_{\theta}(\z_t, t)
    \rVert^2.
\end{equation*}
We further parameterize $\boldsymbol{\mu}_{\theta}(\z_t, t) := \tilde{\boldsymbol{\mu}}_t(\z_t, \zhat(\z_t, t))$, where $\zhat(\z_t, t)$ is the model prediction of the original data $\z_0$. Consequently,
\begin{align*}
    \small
    \loss_{t - 1}
    &= \frac{1}{2 \sigma_t^2} 
    \lVert
        \xi_t \z_0 + \lambda_t \z_t
        - (\xi_t \zhat(\z_t, t) + \lambda_t \z_t)
    \rVert^2 \\
    &\quad + C_3 \\
    &= \frac{\lambda_t^2}{2 \sigma_t^2} 
    \lVert \z_0 - \zhat(\z_t, t)) \rVert^2 + C_3.
\end{align*}

For the last term,
\begin{equation*}
    \small
    -\log p_\theta(\z_0 | \z_1)
    = \frac{1}{2 \sigma_1^2} 
    \lVert
        \z_0
        - \tilde{\boldsymbol{\mu}}_1(\z_1, \zhat(\z_1, 1))
    \rVert^2 + C_4.
\end{equation*}
Noting that
\begin{equation*}
    \left\{
        \begin{aligned}
            \xi_1 &=
            \frac{\sqrt{1} (1 - \alpha_1)}
            {1 - \alpha_1} = 1 \\
            \lambda_1 &=
            \frac{\sqrt{\alpha_1} (1 - 1)}
            {1 - \alpha_1} = 0
        \end{aligned}
    \right.,
\end{equation*}
the term can be converted as
\begin{align*}
    -\log p_\theta(\z_0 | \z_1)
    &= \frac{\lambda_1}{2 \sigma_1^2} 
    \lVert \z_0 - \zhat(\z_1, 1) \rVert^2 + C_4 \\
    &= \left. \loss_{t - 1} \right|_{t = 1} + C_4',
\end{align*}
which can be combined with the sum of the KL terms.

Finally, ignoring constant terms and weight factors, the training objective becomes
\begin{equation*}
    \lvlb = \mathbb{E}_{\z_0, \z_t, t} \left [
        \lVert \zhat(\z_t, t) - \z_0 \rVert^2
    \right ].
\end{equation*}

\section{Additional Experimental Results}

\subsection{More Metrics in Translation}
\label{sec:metrics}

We evaluate the COMET~\citep{rei2020comet} score of Difformer as well as baselines on the translation task in \cref{tab:comet}. Combined with BLEU, these results also confirm the performance of Difformer, and its comparability with autoregressive models.

\begin{table}[t]
    \small
    \centering
    \begin{tabular}{l *{3}{c}}
        \toprule
        \multirow{2}{*}{\textbf{Models}} &
        \textbf{WMT14 En-De} & \textbf{IWSLT14 De-En} \\
        & \textbf{COMET} & \textbf{COMET} \\
        \midrule
        
        Transformer & $0.8286$ & $0.7894$ \\
        CMLM & $0.8226$ & $0.7736$ \\
        \midrule

        Difformer & $0.8257$  & $0.7875$ \\
        \bottomrule
    \end{tabular}
    \caption{The COMET scores of Difformer and baselines. All results are reported by our implementation.}
    \label{tab:comet}
\end{table}

\subsection{Comparison with More Baselines}
\label{sec:baselines}

As our research in this work focuses on optimization challenges of embedding diffusion models, we mainly compare Difformer with existing diffusion-based models. To evaluate the promising performance of Difformer, we extend our comparisons to include results from established traditional models, such as mBART~\citep{liu2020multilingual}, Levenshtein Transformer~\citep{gu2019levenshtein}, DisCo~\citep{kasai2020non}, Fully NAT~\citep{gu2021fully}, and DA-Transformer~\citep{huang2022directed}. The results are presented in \cref{tab:more_baselines}. It is worth noting that mBART is a pre-trained language model, which employs significantly larger datasets in training, and it is more than $10$ times larger in terms of parameter number. Through the results, the performance of Difformer is still competitive even against more recent and stronger baselines.

\begin{table}[t]
    \small
    \centering
    
    \begin{tabular}{lcc}
        \toprule
        \textbf{Models} & $b$ & \textbf{BLEU} \\
        
        \midrule
        mBART & $5$ & $30.50$ \\
        
        \midrule
        Levenshtein Transformer & $5$ & $27.27$ \\
        DisCo & $5$ & $27.34$ \\
        Fully NAT & $1$ & $27.49$ \\
        DA-Transformer & $200$ & $27.78$ \\
        
        \midrule
        \textbf{Difformer} & $5$ & $27.61$ \\
    \bottomrule
    
    \end{tabular}
    \caption{BLEU scores on the WMT14 En-De dataset. All results are as reported in their paper.}
    \label{tab:more_baselines}
\end{table}

\subsection{Study of Frozen Embeddings}
\label{sec:frozen_emb}

\cref{sec:collapse} discusses the collapse problem arising from learnable embeddings and introduces the anchor loss to prevent collapse and ensure stability of training. Alternatively, replacing the embedding space with a pre-trained one is also a solution to address this problem, and our preliminary experiments explore this possibility. The corresponding experimental results can be found in~\cref{tab:frozen_emb}.
From the table, the embeddings from pre-trained CMLM alleviate the collapse problem notably, and the performance obtained by each setting is highly related to $\ani$, which echos our findings in \cref{sec:collapse}. Moreover, our proposed method exhibits a substantial improvement compared with the model with pre-trained embeddings. We attribute this to the possibility that the embeddings from traditional models are suboptimal for diffusion models.

\begin{table}[t]
    \small
    \centering
    
    \begin{tabular}{lcc}
        \toprule
        \textbf{Setting} & $\ani$ & BLEU \\
        \midrule

        $\lvlb$ & $0.99$ & $0.07$ \\
        $\lvanilla$ & $0.32$ & $27.89$ \\
        $\lours$ & $0.03$ & $\textbf{34.48}$ \\
        $\lours$ w/ fixed CMLM embeddings & $0.02$ & $30.14$ \\
        \bottomrule
    \end{tabular}
    \caption{The anisotropy score and performance of different settings on the IWSLT14 De-En dataset with the \emph{linear} schedule.}
    \label{tab:frozen_emb}
\end{table}

\begin{table*}[t]
    \small
    \centering

    \begin{tabular}{c cc cc cc}
        \toprule
        \multirow{2}{*}{$\dgsmax$} &
        \multicolumn{2}{c}{\textbf{\emph{Linear}}} &
        \multicolumn{2}{c}{\textbf{\emph{Cosine}}} &
        \multicolumn{2}{c}{\textbf{\emph{Sqrt}}} \\

        \cmidrule(lr){2-3} \cmidrule(lr){4-5} \cmidrule(lr){6-7} &
        
        \textbf{RF} & \textbf{BLEU} &
        \textbf{RF} & \textbf{BLEU} &
        \textbf{RF} & \textbf{BLEU} \\
        \midrule
        
        $0.05$ &
        $21.0$ & $31.99$ &
        $41.0$ & $31.64$ &
        $7.0$ & $33.88$ \\
        
        $0.15$ &
        $6.0$ & $33.09$ &
        $12.5$ & $33.36$ &
        $4.0$ & $34.48$ \\
        
        $0.50$ &
        - & - &
        $3.0$ & $33.01$ &
        $2.0$ & $32.54$ \\

        - &
        $1.0$ & $32.21$ &
        $1.0$ & $26.61$ &
        $1.0$ & $22.70$ \\
        \bottomrule
    \end{tabular}
    \caption{The performance with varying $\dgsmax$ and noise schedule. Where RF stands for the rescaling factor. Refer to \cref{tab:dgs}, $\dgs$ of \emph{linear} is less than $0.50$, thus the result of \emph{linear} with $\dgsmax = 0.50$ is empty. The last row represents the model without noise rescaling. Results are obtained on the IWSLT14 De-En dataset with $b = 10$.}
    \label{tab:dgsmax_schedules}
\end{table*}

\subsection{Study of $\dgsmax$ with Different Schedules}
\label{sec:dgsmax_schedules}

We further provide more results with different values of $\dgsmax$ and schedules in \cref{tab:dgsmax_schedules}. From the table, we can notice that the degeneration problem exists widely in noise schedules, and the proposed noise rescaling framework enhances the performance of schedules consistently. Furthermore, the rescaling factor of \emph{sqrt} is more sensitive to $\dgsmax$.

\subsection{Study of the Dimension of the Embeddings}
\label{sec:dim_emb}

Prior works mainly use an embedding dimension of $128$~\citep{li2022diffusion, gong2022diffuseq, yuan2023seqdiffuseq}, and we also find the model is quite hard to work with a higher embedding dimension like $256$ or $512$. Intuitively, higher dimensional embedding space is sparser, and embeddings require more noise to diffuse from their nearest neighbor region. Therefore, the model encounters more severe degeneration in this space if provided with insufficient noise. From \cref{tab:dim_emb}, a larger rescaling factor is demanded to reach the same $\dgsmax$ in a higher dimensional embedding space. On the other hand, a low-dimensional embedding space may result in a limited capacity for representation. With noise rescaling, models with different embedding dimensions work successfully and achieve similar results, showing the robustness of the proposed noise rescaling framework. Additionally, the noise rescaling framework augments the scalability of embedding diffusion models and actualizes the potential of application on large-scale datasets and tasks.

\begin{table}[t]
    \small
    \centering

    \begin{tabular}{ccc}
        \toprule
        \textbf{$d$} & \textbf{Rescaling Factor} & \textbf{BLEU} \\
        \midrule
        
        $64$ & $2.5$ & $32.11$ \\
        $128$ & $4.0$ & $34.48$ \\
        $256$ & $6.0$ & $34.41$ \\
        $512$ & $8.5$ & $34.33$ \\
        \bottomrule
    \end{tabular}
    \caption{The performance with varying dimensions of the embedding space. Results are obtained on the IWSLT14 De-En dataset with $b = 10$ and $\dgsmax = 0.15$.}
    \label{tab:dim_emb}
\end{table}

\subsection{Study of Beam Size}
\label{sec:beam}

We study the influence of the 2D parallel decoding hyper-parameters, \ie, the length beam size $b_1$ and noise beam size $b_2$ described in \cref{sec:2d}. As shown in \cref{tab:beam}, we find that length and noise beams both boost the generation quality and are complementary to each other. The length beam brings more significant improvements, while a sufficiently large $b = b_1 \times b_2$ leads to the saturation of the BLEU score.

\begin{table}[t]
    \small
    \centering
    
    \begin{tabular}{ccc}
        \toprule
        $b_1$ & $b_2$ & \textbf{BLEU} \\
        \midrule
        
        $1$ & $1$ & $32.91$ \\
        \midrule
        
        $1$ & $9$ & $33.53$ \\
        $3$ & $3$ & $34.33$ \\
        $5$ & $2$ & $34.43$ \\
        $9$ & $1$ & $34.48$ \\
        \midrule

        $5$ & $4$ & $34.52$ \\
        $7$ & $3$ & $34.48$ \\
        $9$ & $2$ & $34.44$ \\
        \midrule

        $10$ & $5$ & $34.52$ \\
        \bottomrule
    \end{tabular}
    \caption{The performance with varying $b_1$ and $b_2$. Results are obtained on the IWSLT14 De-En dataset.}
    \label{tab:beam}
\end{table}

\subsection{Noise Rescaling in Sampling}

Since noise rescaling is a technique to alleviate a training problem, it is not applied in sampling. We study the sampling quality when applying the noise rescaling in decoding steps. As illustrated in \cref{tab:nr_sampling}, the noise rescaling in sampling is harmful to sampling quality due to no requirement of large noise in the reverse process.

\begin{table}[t]
    \small
    \centering

    \begin{tabular}{cc}
        \toprule
        \textbf{NR in Sampling} & \textbf{BLEU} \\
        \midrule

        & $34.48$ \\
        \checkmark & $33.51$ \\
        \bottomrule
    \end{tabular}
    \caption{The BLEU score with noise rescaling~($\dgsmax = 0.15$) in sampling. Results are obtained on the IWSLT14 De-En dataset with $b = 10$.}
    \label{tab:nr_sampling}
\end{table}

\begin{table*}[t]
    \small
    \centering

    \begin{tabular}{cccc}
        \toprule
        \textbf{$\dgsmax$} & \textbf{Variance Preserving} & \textbf{Rescaling Factor} & \textbf{BLEU} \\
        \midrule
        
        \multirow{2}{*}{$0.05$} && $7.0$ & $33.88$ \\
        & \checkmark & $9.5$ & $33.80$ \\
        \midrule
        
        \multirow{2}{*}{$0.15$} && $4.0$ & $34.48$ \\
        & \checkmark & $4.5$ & $33.99$ \\
        \midrule
        
        \multirow{2}{*}{$0.50$} && $2.0$ & $32.54$ \\
        & \checkmark & $2.0$ & $31.77$ \\
        \midrule
        
        $0.77$ & - & $1.0$ & $22.70$ \\
        \bottomrule
    \end{tabular}
    \caption{The performance with varying $\dgsmax$ and variance-preserving rescaling factor. Results are obtained on the IWSLT14 De-En dataset with $b = 10$.}
    \label{tab:vprf}
\end{table*}

\subsection{Variance-Preserving Rescaling Factor}
\label{sec:vprf}

The variance-preserving variant of the rescaling factor can be defined as
\begin{equation}
    \label{equ:vprf}
    \left\{
        \begin{aligned}
            \alphabar_t' &= \frac{\alphabar_t}{\alphabar_t + F^2 \betabar_t} \\
            \betabar_t' &= \frac{F^2 \betabar_t}{\alphabar_t + F^2 \betabar_t}
        \end{aligned}
    \right.,
\end{equation}
where $F$ is the rescaling factor. We can derive that $\alphabar_t' + \betabar_t' = 1$ and the signal-to-noise ratio of the rescaled schedule satisfies $\mathrm{SNR}_t' = \nicefrac{\mathrm{SNR}_t}{F^2}$, where $\mathrm{SNR}_t = \nicefrac{\alphabar_t}{\betabar_t}$. The VP rescaling factor performs similarly to the original version, which is listed in the \cref{tab:vprf}.

\subsection{Controllable Generation}
\label{sec:control}

Due to the distinctive iterative generation process, one of the advantages of diffusion models is that they support flexible and fine-grained control over the outputs, such as syntax tree, length, prefix, and suffix~\citep{li2022diffusion}. We validate the length conditioning ability for Difformer in \cref{tab:len_control}. Through the cases, the length of generations is well-matched with the condition, highlighting the controllability of Difformer.

\begin{table}[t]
    \small
    \centering
    
    \begin{tabular}{cm{0.7\columnwidth}}
        \toprule
        \textbf{Length} & \multicolumn{1}{c}{\textbf{Generations}} \\
        \midrule

        $5$ & \texttt{our imagination is even reality.} \\
        $10$ & \texttt{our imagination is a force that can even create reality.} \\
        $15$ & \texttt{our imagination, in fact, is a force that can even create a reality.} \\

        \midrule
        \textbf{Target} &
        \texttt{imagination is a force that can actually manifest a reality.} \\
        \bottomrule
    \end{tabular}
    \caption{Cases when applying length control.}
    \label{tab:len_control}
\end{table}

\subsection{Case Study}

To qualitatively analyze the dynamics of the generation quality at the instance level, we select several representative cases in \cref{tab:cases}.
The generation results in \cref{tab:cases} illustrate the characteristics of Difformer. Specifically, after the first few steps, the model is able to transform random words into noised but human-readable sentences, and with the 
reverse process progressing, these sentences are refined gradually. However, as the process enters the final steps, the model at these timesteps suffers from the degeneration problem and loses the ability to improve the outputs, even corrupts previously correct words. This dynamics is also reflected in \cref{fig:dynamics}, and emphasizes the necessity of noise rescaling and early stopping techniques.
On the other hand, Difformer achieves comparable quality with the autoregressive baseline, and generates more coherent and consistent sentences compared with traditional non-autoregressive models.

\begin{algorithm}[t]
    \caption{Search $F$}
    \label{alg:f}
    
    \textbf{Input}: Noise schedule $\alphabar, \betabar$, threshold of degeneration score $\dgsmax$, search interval $\Delta F$ \\
    \textbf{Output}: $F$

    \begin{algorithmic}[1]
        \STATE $F \gets 1$
        \WHILE{\TRUE}
            \STATE Rescale $\alphabar'$ and $\betabar'$ from $\alphabar$ and $\betabar$ using $F$ according to \cref{equ:rf} or \cref{equ:vprf}
            \STATE $\dgs \gets \dgs \left( \alphabar', \betabar' \right)$
            \IF{$\dgs \le \dgsmax$}
                \RETURN $F$
            \ENDIF
            \STATE $F \gets F + \Delta F$
        \ENDWHILE
        
    \end{algorithmic}
\end{algorithm}

\begin{algorithm}[t]
    \caption{Compute $\dgs$}
    \label{alg:dgs}
    \textbf{Input}: Noise schedule $\alphabar, \betabar$, embeddings $\eset$, timestep set $\mathcal{T}$, repeat times $N$ \\
    \textbf{Output}: $\dgs$

    \begin{algorithmic}[1]
        \STATE $\dgs \gets 0$
        \FORALL{$t \in \mathcal{T}$}
            \STATE $\dgs_t \gets 0$
            \FORALL{$\zvec_0 \in \eset$}
                \FOR{$i \gets 1$ \TO $N$}
                    \STATE Sample $\zvec_t \sim \mathcal{N} \left( \sqrt{\alphabar_t} \zvec_0, \betabar_t \mathbf{I} \right)$
                    \IF {$\fdg(\zvec_t) = \zvec_0$}
                        \STATE $\dgs_t \gets \dgs_t + 1$
                    \ENDIF
                \ENDFOR
            \ENDFOR
            \STATE $\dgs_t \gets \dgs_t / (N \times V)$
            \STATE $\dgs \gets \dgs + \dgs_t$
            \ENDFOR
        \STATE $\dgs \gets \dgs / \lvert \mathcal{T} \rvert$
        \RETURN $\dgs$
    \end{algorithmic}
\end{algorithm}

\section{Search of Rescaling Factor}
\label{sec:search_rf}

To decide $F$, we can perform either brute-force or binary search. Without loss of generality, the approach of brute-force search is presented in \cref{alg:f}.
For the computation of $\dgs$, we firstly regard $\fdg$ as an equivalent nearest neighbor classifier, then utilize a Monte Carlo method to estimate $\dgs_t$ at some of the timesteps, and finally compute the average of all values of $\dgs_t$. \cref{alg:dgs} illustrates this process.
Since our theorem is based on the assumption that $\emat \sim \mathcal{N}_{d \times V} (\mathbf{0}, \sigma_e \mathbf{I})$, we initialize it with a normal distribution, and the computation of $\dgs$ can be independent of real embeddings and approximated before training.
For the convenience of future research, we provide a pre-computed function table of $\dgs$ and the corresponding rescaling factors in \cref{tab:rf}.

\begin{table*}[t]
    \small
    \centering

    \begin{tabular}{*{10}{c}}
        \toprule
        \multirow{2}{*}{$\dgsmax$} &
        \multicolumn{3}{c}{\textbf{\emph{Linear}}} &
        \multicolumn{3}{c}{\textbf{\emph{Cosine}}} &
        \multicolumn{3}{c}{\textbf{\emph{Sqrt}}} \\

        \cmidrule(lr){2-4}
        \cmidrule(lr){5-7}
        \cmidrule(lr){8-10} &
        
        $10$K & $20$K & $40$K &
        $10$K & $20$K & $40$K &
        $10$K & $20$K & $40$K \\
        \midrule
        
        $0.05$ &
        $21.0$ & $20.0$ & $19.0$ &
        $41.0$ & $38.5$ & $37$ &
        $7.0$ & $6.5$ & $6.0$ \\
        
        $0.10$ &
        $9.5$ & $9.0$ & $9.0$ &
        $19.0$ & $18.5$ & $18.0$ &
        $5.0$ & $4.5$ & $4.5$ \\
        
        $0.15$ &
        $6.0$ & $5.5$ & $5.5$ &
        $12.5$ & $12.0$ & $11.5$ &
        $4.0$ & $4.0$ & $3.5$ \\

        $0.20$ &
        $4.0$ & $4.0$ & $4.0$ &
        $9.0$ & $9.0$ & $8.5$ &
        $3.5$ & $3.5$ & $3.0$ \\

        $0.30$ &
        $2.5$ & $2.5$ & $2.5$ &
        $6.0$ & $5.5$ & $5.5$ &
        $2.5$ & $2.5$ & $2.5$ \\

        $0.50$ &
        - & - & - &
        $3.0$ & $2.5$ & $2.5$ &
        $2.0$ & $2.0$ & $1.5$ \\
        \bottomrule
    \end{tabular}
    \caption{The $\dgsmax$ and corresponding rescaling factors with different vocabulary sizes and noise schedules. We perform the search in the $128$-dimensional embedding space at the interval of $0.5$, and $\sigma_e = 1$.}
    \label{tab:rf}
\end{table*}

\begin{table*}[t]
    \small
    \centering

    \begin{tabular}{l*{7}{C{4em}}}
        \toprule
        \textbf{Splits} &
        \textbf{WMT14 En-De} &
        \textbf{WMT16 En-Ro} &
        \textbf{IWSLT14 De-En} &
        \textbf{Gigaword} &
        \textbf{QQP} &
        \textbf{Wiki-Auto} &
        \textbf{QT} \\
        \midrule
        
        Training &
        $4,500,966$ & $608,319$ & $160,215$ & $3,803,957$ &
        $144,715$ & $677,751$ & $116,953$ \\

        Validation &
        $3,000$ & $1,999$ & $7,282$ & $189,651$ &
        $2,048$ & $2,048$ & $2,048$ \\

        Test &
        $3,003$ & $1,999$ & $6,750$ & $1,951$ &
        $2,500$ & $5,000$ & $10,000$ \\
        \bottomrule
    \end{tabular}
    \caption{The dataset splits used in our experiments.}
    \label{tab:dataset}
\end{table*}

\section{Experimental Settings}
\label{sec:settings}

We build our model based on Transformer~\citep{vaswani2017attention} and use \texttt{transformer-iwslt-de-en} config for the IWSLT dataset, \texttt{transformer-base} config for WMT and summarization datasets. For other datasets, to conduct fair comparisons with DiffuSeq, we set the model dimension to $768$ and the feed-forward intermediate dimension to $3072$. We tokenize sentences and segment each token into subwords by Byte-Pair Encoding~\citep{sennrich2016neural}.
The training process takes nearly one day on $8$ NVIDIA V100 GPUs for the WMT datasets and the Gigaword dataset, while nearly $12$ hours on one NVIDIA V100 GPU for the other datasets.
In inference, we downsample the diffusion step to $20$, \ie, $K = 20$, and stop the decoding process $5$ steps earlier, which is much faster than previous works~\citep{li2022diffusion, gong2022diffuseq} while maintaining the performance. Each reported result is the average of $3$ runs. The dataset splits we used are listed in \cref{tab:dataset}. All datasets can be used for research purposes. The detailed hyper-parameters are listed in \cref{tab:settings}.

\begin{table*}[t]
    \small
    \centering

    \begin{tabular}{l*{7}{C{4em}}}
        \toprule
        \textbf{Hyper-parameters} &
        \textbf{WMT14 En-De} &
        \textbf{WMT16 En-Ro} &
        \textbf{IWSLT14 De-En} &
        \textbf{Gigaword} &
        \textbf{QQP} &
        \textbf{Wiki-Auto} &
        \textbf{QT} \\
        \midrule
        
        \textbf{Architecture} \\
        
        $d_{\mathrm{model}}$ &
        $512$ & $512$ & $512$ & $512$ &
        $768$ & $768$ & $768$ \\

        $d_{\mathrm{emb}}$ &
        $128$ & $128$ & $128$ & $128$ &
        $128$ & $128$ & $128$ \\

        $d_{\mathrm{ffn}}$ &
        $2048$ & $2048$ & $1024$ & $2048$ &
        $3072$ & $3072$ & $3072$ \\

        Heads &
        $8$ & $8$ & $4$ & $8$ &
        $12$ & $12$ & $12$ \\

        Encoder Layers &
        $6$ & $6$ & $6$ & $6$ &
        $6$ & $6$ & $6$ \\

        Decoder Layers &
        $6$ & $6$ & $6$ & $6$ &
        $6$ & $6$ & $6$ \\

        Activation &
        ReLU & ReLU & ReLU & ReLU &
        ReLU & ReLU & ReLU \\
        \midrule

        \textbf{Diffusion} \\

        Steps &
        $2000$ & $2000$ & $2000$ & $2000$ &
        $2000$ & $2000$ & $2000$ \\

        Schedule &
        \emph{sqrt} & \emph{sqrt} & \emph{sqrt} & \emph{sqrt} &
        \emph{sqrt} & \emph{sqrt} & \emph{sqrt} \\

        $\dgsmax$ &
        $0.15$ & $0.15$ & $0.15$ & $0.15$ &
        $0.15$ & $0.15$ & $0.20$ \\

        Self-Conditioning &
        \checkmark & \checkmark & \checkmark & \checkmark &
        \checkmark & \checkmark & \checkmark \\
        \midrule

        \textbf{Training} \\
        Steps &
        $300$K & $300$K & $300$K & $300$K &
        $50$K & $30$K & $100$K \\

        Batch Size (Tokens) &
        $64$K & $64$K & $8$K & $64$K &
        $8$K & $64$K & $64$K \\

        Optimizer &
        AdamW & AdamW & AdamW & AdamW &
        AdamW & AdamW & AdamW \\

        Adam $\beta$ &
        $(0.9, 0.98)$ & $(0.9, 0.98)$ &
        $(0.9, 0.98)$ & $(0.9, 0.98)$ &
        $(0.9, 0.98)$ & $(0.9, 0.98)$ &
        $(0.9, 0.98)$ \\

        Weight Decay &
        $0.01$ & $0.01$ & $0.01$ & $0.01$ &
        $0.01$ & $0.01$ & $0.01$ \\

        Learning Rate &
        $5 \times 10^{-4}$ & $5 \times 10^{-4}$ &
        $5 \times 10^{-4}$ & $5 \times 10^{-4}$ &
        $5 \times 10^{-4}$ & $5 \times 10^{-4}$ &
        $3 \times 10^{-4}$ \\
        
        Warmup &
        $10$K & $10$K & $10$K & $10$K &
        $10$K & $10$K & $5$K \\

        Clip Gradient &
        $1.0$ & $1.0$ & $1.0$ & $1.0$ &
        $1.0$ & $1.0$ & $1.0$ \\

        Dropout &
        $0.1$ & $0.1$ & $0.3$ & $0.1$ &
        $0.1$ & $0.1$ & $0.1$ \\

        Length Predict Factor &
        $0.1$ & $0.1$ & $0.1$ & $0.1$ &
        $0.1$ & $0.1$ & $0.1$ \\

        Label Smoothing &
        $0.1$ & $0.1$ & $0.1$ & $0.1$ &
        $0.1$ & $0.1$ & $0.1$ \\
        \midrule

        \textbf{Inference} \\
        Steps &
        $20$ & $20$ & $20$ & $20$ &
        $20$ & $20$ & $20$ \\

        Early Stopping &
        $5$ & $5$ & $5$ & $5$ &
        $5$ & $5$ & $5$ \\
        
        \bottomrule
    \end{tabular}
    \caption{The model architectures and hyper-parameters used in our experiments.}
    \label{tab:settings}
\end{table*}

\begin{table*}[t]
    \small
    \centering
    
    \begin{tabular}{rm{0.8\textwidth}}
        \toprule
        \multicolumn{2}{c}{\textbf{Case 1}} \\
        \toprule
        
        Source &
        \texttt{ich denke , dass es schwer wird, sie zu treffen, aber ich denke, es ist es auf jeden fall wert, einige wirklich bekannte marken anzusprechen.} \\

        Target &
        \texttt{i think that you're going to have a hard time meeting with them, but i think it's certainly worth pursuing a couple big, really obvious brands.} \\

        \midrule
        \textbf{Progress} & \multicolumn{1}{c}{\textbf{Generation}} \\
        \midrule

        $0\%$ &
        \texttt{i think it's going to meet you, but i i it it definitely worth talking sing some really known brands.} \\
        
        $25\%$ &
        \texttt{i think it's hard to meet you, but i think it's definitely worth mating some really known brands.} \\
        
        $50\%$ &
        \texttt{i think it's hard to meet you, but i think it's definitely worth approaching some really known brands.} \\
        
        $75\%$ &
        \texttt{i think it's hard to meet you, but i think it's definitely worth approaching some really known brands.} \\

        $100\%$ &
        \texttt{i think it's hard to meet you, but i think it's definitely worth approaching some really known brks.} \\

        \midrule
        Transformer &
        \texttt{i think it's hard to meet you, but i think it's worth addressing some really familiar brands, i think.} \\

        CMLM &
        \texttt{i think it's hard to meet you, but i think it's worth saying some really known brands.} \\

        \toprule
        \multicolumn{2}{c}{\textbf{Case 2}} \\
        \toprule

        Source &
        \texttt{und wenn wir mit einem körper konfrontiert sind, der für uns tatsächlich etwas sehr anderes darstellt, verwirrt uns das in hinblick auf diese kategorisierungen.} \\

        Target &
        \texttt{and when we're faced with a body that actually presents us something quite different, it startles us in terms of those categorizations.} \\

        \midrule
        \textbf{Progress} & \multicolumn{1}{c}{\textbf{Generation}} \\
        \midrule

        $0\%$ &
        \texttt{and when we're confronted with a body which is actually something very different for us, we're confused in in terms of these categorization.} \\
        
        $25\%$ &
        \texttt{and when we're confronted with a body that is actually something very different for us, we're confused it in terms of categcategorization.} \\
        
        $50\%$ &
        \texttt{and when we're confronted with a body that's actually something very different for us, we're confusing it in terms of these categorization.} \\
        
        $75\%$ &
        \texttt{and when we're confronted with a body that's actually something very different for us, we're confusing it in terms of these categorization.} \\

        $100\%$ &
        \texttt{and when we're confronted with a body that's actually something very different for us, we're confusing it in terms of these categorizes.} \\

        \midrule
        Transformer &
        \texttt{and when we're faced with a body, which actually represents something very different for us, it confuses us in terms of these categorization.} \\

        CMLM &
        \texttt{and when we're faced with a body that actually represents something very different for us, it confuses us in terms of these categorization.} \\

        \toprule
        \multicolumn{2}{c}{\textbf{Case 3}} \\
        \toprule

        Source &
        \texttt{um also das blinken zu beschleunigen oder zu verlangsamen, drehen sie einfach an diesem knopf und er macht den impuls schneller oder langsamer.} \\

        Target &
        \texttt{so to make this blink faster or slower, you would just turn this knob and basically make it pulse faster or slower.} \\

        \midrule
        \textbf{Progress} & \multicolumn{1}{c}{\textbf{Generation}} \\
        \midrule

        $0\%$ &
        \texttt{so to accelerate or flck slow slow down, , just turn on buttbuttand it makes pulse faster or slow wer.} \\
        
        $25\%$ &
        \texttt{so to accelerate the ck or slow it down, just turn on this button and makes the impulse faster or slow down.} \\
        
        $50\%$ &
        \texttt{so to accelerate the blind or slow it down, just turn on this button and makes the impulse faster or slow down.} \\
        
        $75\%$ &
        \texttt{so to accelerate the blind or slow it down, just turn on this button and makes the impulse faster or slower.} \\

        $100\%$ &
        \texttt{so to accelerate the blind or slow it down, just turn on this button and makes the impulse faster or slower.} \\

        \midrule
        Transformer &
        \texttt{so to slow the blind up or slow the blind down, you just turn that button, and it makes the pulse faster or slower.} \\

        CMLM &
        \texttt{so to speed your blind up or slow down, just turn on that button and it makes the pulse faster or slow.} \\
        
        \bottomrule
    \end{tabular}
    \caption{Cases of intermediate generation results during the whole generation process, compared with baselines.}
    \label{tab:cases}
\end{table*}

\end{document}